\documentclass{article} 
\usepackage{iclr2027_conference,times}
\iclrfinalcopy

\usepackage{amsmath,amsfonts,bm}

\def\eqref#1{equation~\ref{#1}}

\def\1{\bm{1}}

\DeclareMathAlphabet{\mathsfit}{\encodingdefault}{\sfdefault}{m}{sl}
\SetMathAlphabet{\mathsfit}{bold}{\encodingdefault}{\sfdefault}{bx}{n}

\usepackage{microtype}
\usepackage{xcolor}
\usepackage{xspace}
\usepackage{hyperref}
\usepackage{cleveref}
\usepackage{amsmath}
\usepackage{microtype}
\usepackage{xcolor}
\usepackage{booktabs}
\usepackage{tabularx}
\usepackage{array}
\usepackage{float}
\usepackage{placeins}
\usepackage{colortbl}
\usepackage{graphicx}
\usepackage{wrapfig}
\usepackage{capt-of}
\definecolor{figgray}{HTML}{666A70}
\definecolor{gainFill}{HTML}{E5F1E7}
\definecolor{whitesmoke}{HTML}{F5F5F5}

\newcommand{\sys}[0]{CacheBack\xspace}

\hypersetup{
  hidelinks,
  pdftitle={Receiver-Conditioned Latent Communication gives 94\% CacheBack},
  pdfauthor={Maximillian Rossi, Prajwal Raghunath, Haoqing Xuan, Yusen Zhang, Eugene Wu},
  pdfsubject={Preprint}
}

\title{Receiver-Conditioned Latent Communication gives 94\% CacheBack}

\author{
\normalfont Maximillian Rossi\textsuperscript{1} \quad Prajwal Raghunath\textsuperscript{1} \quad Haoqing Xuan\textsuperscript{2}\thanks{This work does not represent the views of Amazon Web Services.} \\
Yusen Zhang\textsuperscript{1} \quad Eugene Wu\textsuperscript{1} \\
\textsuperscript{1}\raisebox{-0.15\height}{\includegraphics[height=1.1em]{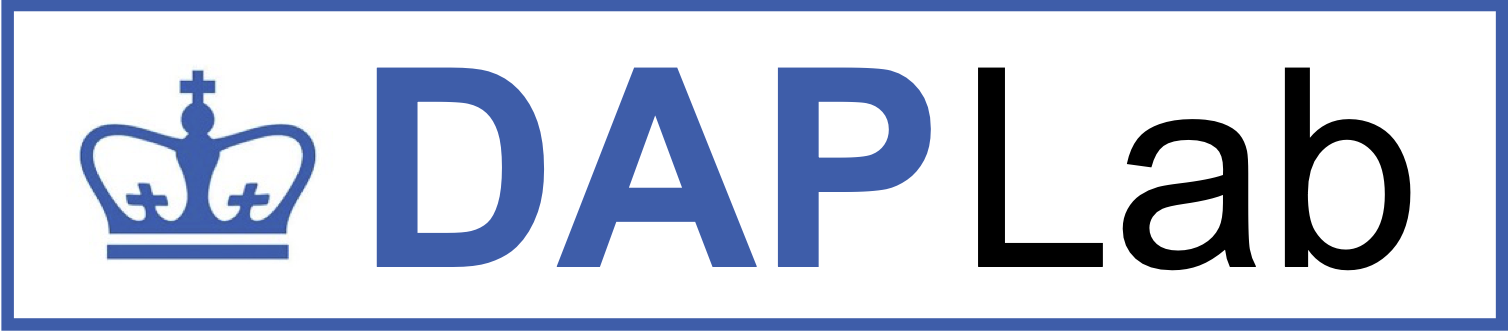}}\, Columbia University \quad \textsuperscript{2}Amazon Web Services \\
\href{mailto:mfr2178@columbia.edu}{\texttt{mfr2178@columbia.edu}}
}

\begin{document}
\vspace*{-36pt}
\maketitle\vspace{-20pt}
\lhead{Preprint}
\centerline{\textbf{Code: \href{https://github.com/maxr0ssi/rclc}{\texttt{github.com/maxr0ssi/rclc}}}}
\vspace{2pt}
\begin{abstract}
Multi-agent systems distribute large contexts across agents that communicate to solve a task. Text messages are compact but require decoding and may omit evidence the receiving agent needs. Recent latent communication instead transfers KV caches. This avoids text generation and can improve accuracy and latency. However, a full KV cache grows linearly with both the context an individual agent processes, and the number of agents that coordinate together. This raises memory and context costs, often far exceeding available GPU resources and context window sizes.
Our key observation is that agents need only send what the receiving agent requires for its local task---which we call {\it receiver-conditioned communication}. The receiver agent passes the sender a small description of its information needs, which serves to filter and compress the sender agent's KV cache. \textbf{CacheBack} is a simple, robust, training-free instance of {\it receiver conditioning} based on the sender's attention weights. On FanOutQA, CacheBack with Qwen 3 removes 75\% of the state the agent would otherwise receive, improving accuracy by 14.7 percentage points and reducing median task-completion latency by \(3.2\times\) relative to text communication. We show comparable improvements across model families that span dense Transformers, Mamba-attention hybrids, and sliding-window attention.

\end{abstract}
\vspace{-14pt}
\begin{figure}[h]
\centering
\includegraphics[width=\linewidth]{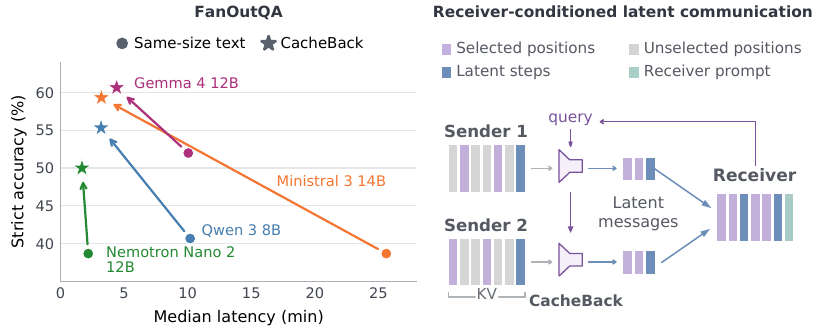}
\caption{Left: highest-accuracy CacheBack setting per family versus same-size text on FanOutQA; $n=50$ concurrent tasks on $8\times$ H100. Right: the receiver query guides selection of sender KV states.}
\label{fig:fanoutqa-overview}
\end{figure}

\vspace{-6pt}
\section{Introduction}
\label{sec:intro}

Agents commonly distribute information and work across models to exploit parallelism, manage context limits, and divide tasks into specialized steps \citep{anthropic2025,zhang2024chain,fourney2024magenticone}. We consider directed acyclic graphs, where nodes (agents) send messages from sender to receiver along directed edges; an interior node may both send and receive messages.  Common patterns include fan-in, where work is partitioned across multiple agents who send their outputs to a single receiver, and a sequential chain, where agents process incoming messages and send to the next.

\newpage
For example, a FanOutQA task asks, ``{\it What is the highest elevation (in feet) of the five largest U.S. states by land mass?}'' \citep{zhu-etal-2024-fanoutqa}. Answering it requires identifying the five states from one Wikipedia article, then finding each state's highest elevation in five additional articles. With four agents, we may give two articles to three of the agents to read, and they send summaries to a receiving agent that combines and answers with Alaska 20,310; Texas 8,751; California 14,505; Montana 12,807; New Mexico 13,161.

While text is the dominant form of communication, discrete tokens are less expressive than continuous hidden representations and are costly to generate \citep{zou2025}. In our FanOutQA evaluation, generating three Qwen 3 8B text messages takes a median 73.2 seconds.\footnote{Measured over the same 50 questions on one NVIDIA H100 80GB. Timing includes agent prefill and text generation but excludes receiver computation and transfer.} Systems manage accuracy--latency trade-offs through shorter messages \citep{chen2025optima} or delegation to smaller models \citep{feng2026graphplanner}.

Recent work proposes to use the KV cache for agent communication. LatentMAS generates latent thoughts by autoregressing over hidden representations rather than decoded tokens: each latent step feeds its final hidden representation back as the next continuous input. After $G$ latent steps, an agent sends its full KV cache, containing its input context and generated latent steps, as the message. Assuming that the receiving agent uses the same model, it prepends this message to its KV cache and immediately continues computing. Across reasoning and code benchmarks, LatentMAS improves average accuracy by $2.8$--$4.6$ percentage points and end-to-end latency by $4.0\times$ in sequential and $4.3\times$ in hierarchical topologies over text communication \citep{zou2025}.

However, communicating full KV caches does not scale to long contexts.  When subagents read different articles, their KV caches differ and must all be sent to the receiver.  Thus, the receiver's input context, the amount of transferred data, the memory pressure, and receiver attention must all grow linearly with the number of subagents and article sizes.  This removes the context-management benefit of partitioning evidence across agents. In our FanOutQA settings with Qwen 3 8B, full-cache transfer leaves insufficient context for receiver generation on every task (Section~\ref{sec:fanoutqa}).

We observe that the full KV cache also transmits information the receiver does not need, and that \textbf{what an agent sends should depend on the receiver's information need}. We argue for a {\it receiver-conditioned communication} protocol, where the receiver shares its information need with its subagents, who use the need to selectively choose the information to send. We instantiate this idea in \textbf{CacheBack}, a simple, robust, and training-free selector for communication between the same models that can reduce the message size from $2\times$ to over $100\times$. 

When compared to text communication between same-size models, \sys improves strict accuracy by 7.3--20.7 percentage points and reduces task-completion latency by $1.3\times$--$8.0\times$ on FanOutQA, and improves  accuracy by 9.3--14.7 percentage points with $2.5\times$--$3.7\times$ speedups on LongBench v2.  Across these benchmarks and four model families spanning dense Transformers, Mamba-attention hybrids, and sliding-window attention, we consistently shift the accuracy--latency Pareto frontier towards faster, more accurate task completions.

To summarize, we make three contributions. First, we introduce \textbf{receiver-conditioned communication}, where what an agent sends depends on the receiver's task needs. Second, we introduce \textbf{CacheBack}, a simple, robust, training-free method for receiver-conditioned latent communication that uses the receiver's query to select positions from the sender's sequence. Third, we show that receiver-conditioned latent communication outperforms same-size text communication when both use the same receiver query, in both accuracy and latency across model families and agent topologies.
\FloatBarrier

\section{Agent Communication}
\label{sec:communication-model}

In a DAG agent topology, a message is sent along an edge from a sender agent $i$ to a receiver agent $j$. Agent $i$ first processes $H_i$ input tokens, which include its source context and task instructions, and then produces a message for the receiver. Under text communication, that message contains $K_i$ generated tokens. The sender decodes those tokens one at a time, and the receiver prefills them as token identifiers.

Latent communication instead sends internal model state. Prior work varies both how that state is produced and what form of it is transmitted. Coconut reasons through continuous hidden states \citep{hao2025coconut}, while LatentMAS generates autoregressive latent steps \citep{zou2025}, which we keep fixed. Communication methods send embeddings, hidden states or activations, or KV state
\citep{pham2024cipher,ramesh2025activations,du2026interlat, zheng2025thoughtcomm,zou2025,fu2026cache,shi2025,liu2026beyond}. Some also learn mappings between sender and receiver representations or change how that state is fused at the receiver \citep{fu2026cache,du2026interlat,kriuk2025qkvcomm, rossi2026lcf}. These methods focus on the representation and transmission of latent state. We study which information from the completed sender state should enter the receiver's message.

For the KV-based communication we study, the sender processes $H_i$ input tokens and then generates $G_i$ latent steps. Its completed sequence therefore contains $T_i = H_i + G_i$ positions, and the message is drawn from this sequence. We call each entry of the sender sequence a position; the sender holds one KV entry per layer for each position. A position from an input token has a token ID, while a position from a latent step has no token ID and exists only as a continuous vector. A selected position is one included in the message. Its representation is how that position is encoded for transmission, either as a token ID or as one row of a continuous representation. In long-context settings where senders process large documents, $K_i \ll T_i$.

LatentMAS transmits KV state for all $T_i$ positions at every layer
\citep{zou2025}. KVComm reduces the transmitted state by sharing only selected
layers, but retains all $T_i$ positions within those layers
\citep{shi2025}. Both therefore fix the message content to the sender's entire
sequence while changing how much state is sent for each position.

\subsection{Communication Costs}
\label{sec:communication-costs}

Full-cache communication sends the KV state for every sender-sequence position at every communicated layer; this costs 144\,KiB per position for Qwen 3 8B in bfloat16.

Compression and quantization reduce the message size but not the number of positions the receiver must load.   Even with free transfer costs, the receiver loads all \(T_i\) positions from each agent, in addition to the \(H_R\) positions in the receiver's own prompt.   Thus, \(M\) subagents would consume a receiver's context size of:
\begin{equation}
H_R^{\text{text}} = H_R + \sum\nolimits_{i=1}^{M} K_i,
\qquad
H_R^{\text{full}} = H_R + \sum\nolimits_{i=1}^{M} T_i.
\label{eq:context-reexpansion}
\end{equation}

\begin{figure}[h]
\centering
\includegraphics[width=.8\linewidth]{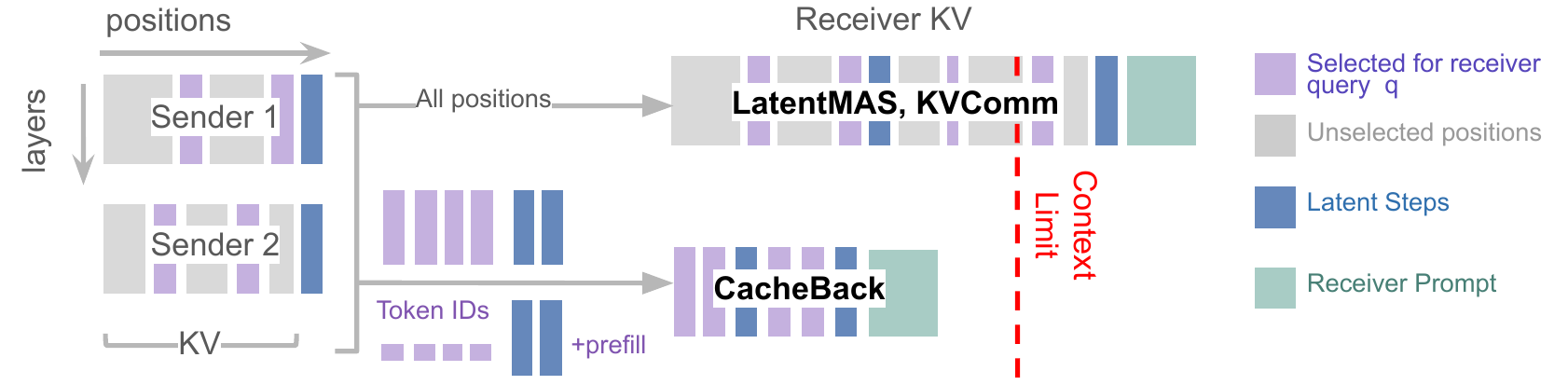}
\caption{Receiver-conditioned communication limits context accumulation. Full-cache messages accumulate all agents' positions at the receiver. Receiver-conditioned messages retain selected source and latent positions, represented as token IDs and continuous inputs for receiver prefill.}
\label{fig:communication-overview}

\end{figure}

We call this context re-expansion. Although partitioning a long input (i.e., many documents) across subagents reduces the context processed per subagent, sending their full states still transfers considerable data over the network, accumulates the contexts at the receiver, consumes the receiver's KV memory, and degrades attention's effectiveness.  In fact, our experiments find that the receiver often does not have enough context for its own prompt.  We illustrate this in Figure~\ref{fig:communication-overview}.

\subsection{The Need for Receiver-Conditioned Selection}

At minimum, context re-expansion must leave room for the receiver's prompt and output. Let $B_i$ bound agent $i$'s selected positions, $O_R$ reserve output positions, and $L_R$ denote the context limit:
\begin{equation}
H_R + {\textstyle\sum_{i=1}^{M}} B_i + O_R \leq L_R.
\label{eq:receiver-capacity}
\end{equation}

Here, $B_i$ bounds positions after re-expansion, and $H_R$ denotes the receiver's prompt length. A position budget limits how much an agent can send, but not which positions to send or how to represent them. Section~\ref{sec:receiver-conditioned-communication} describes how the receiver states what it needs and how the sender uses that query; Appendix~\ref{app:communication-model} formalizes these decisions and their quality--cost objective.

\section{Receiver-Conditioned Communication}
\label{sec:receiver-conditioned-communication}

Receiver-conditioned communication has three parts. First, a receiver query \(q\) states what information the receiver needs. Second, the sender uses \(q\) to choose what information to send. Third, the selected information is represented for transmission to the receiver. Figure~\ref{fig:communication-overview} shows how this reduces the positions sent to the receiver, and the message itself can be further compressed by sending token IDs and continuous vectors (where they do not correspond to vocabulary tokens)  rather than KV contents.

The query \(q\) states what the receiver wants from the sending agent. It is a separate input from the task that prefills the sender's KV, so the sender can service queries from different receivers. In our experiments, \(q\) is the task question. We append it after the sender completes its original prefill and only use $q$ to score sender positions. %

\subsection{Selecting State for the Receiver}
\label{sec:receiver-conditioning-design}
\label{sec:cacheback}

Receiver conditioning does not prescribe how \(q\) should determine message content. While methods could generate new content, transform existing state, or select existing positions, we focus on a simple, training-free setting that selects $\le B_i$ positions using the receiver's query. This design allows us to study the extent to which receiver conditioning helps.

Fortunately, KV-cache compression already provides methods for using a query to decide which source positions to retain, which receiver-conditioned communication sets to the receiver's query $q$.  We study two selector variants: mean query attention, a naive selector based on existing KV-cache compression, and CacheBack, which changes how the query tokens are aggregated.

\subsubsection{Naive selector}
\label{sec:naive-selector}

We build on SnapKV \citep{li2024snapkv}, which scores each earlier position by the attention it receives from the final prompt tokens and keeps the highest-scoring state for the model's own continuation. We put the receiver's query \(q\) in those final positions instead: the sender appends \(q\) and only computes the query tokens against its existing KV cache.

Mean query attention averages the attention these query tokens pay to earlier positions across query tokens and query heads, sums across eligible layers, and pools nearby positions so that they receive similar scores, giving one score \(Q_t\) for each sender-sequence position. Because \(q\) is applied after the sender state is formed, changing the query changes the ranking, and the query can specify a different information need from the task that produced that state. Appendix~\ref{app:cacheback-scoring} describes aggregation, pooling, and eligible layers in more detail.

\noindent\begin{minipage}[t]{0.51\linewidth}
\vspace{0pt}
A limitation of averaging is that positions relevant to one part of the query may only receive strong attention from the few query tokens that refer to them.  Thus, averaging over the whole query dilutes the signal. In Figure~\ref{fig:selection-example}, position A receives strong attention from one query token, while B receives weaker attention from every token, so mean query attention ranks B above A even though A is the position the query asks for. A selector should not penalize a position for being relevant to only part of the query.
\end{minipage}\hfill
\begin{minipage}[t]{0.45\linewidth}
\vspace{0pt}
\centering
\includegraphics[width=\linewidth]{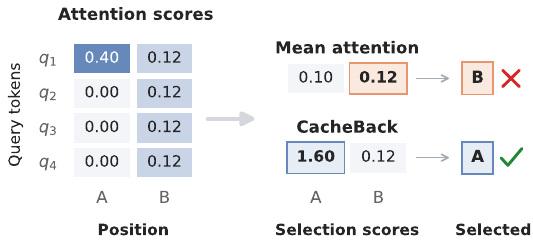}
\captionof{figure}{Averaging can hide query-specific support for \texttt{A} and choose irrelevant token \texttt{B}.}
\label{fig:selection-example}
\end{minipage}
\par
\subsubsection{CacheBack}
\label{sec:cacheback-selector}

CacheBack proposes a simple correction to the averaging problem by reweighting the mean query attention score \(Q_t\) to give additional weight to positions whose attention is concentrated in particular query tokens:
$S_t^{\mathrm{CB}} = Q_t C_t^2.$

Here \(C_t\) compares pooled attention before and after averaging query tokens.
The correction is larger when
a sender position receives concentrated support from part of the query.
Appendix~\ref{app:cacheback-scoring} gives the exact definitions and pooling
order.

We intentionally keep this correction simple and training-free to test whether using $q$ improves communication without introducing a learned module.

\paragraph{State Representation.}
After scoring, we select positions under the message budget \(B_i\)\footnote{The exact span-selection rule is given in Appendix~\ref{app:cacheback-selection}.}. We transmit each selected position and latent step as a continuous row (\textbf{continuous-row encoding}), which is about \(18\times\) smaller per position than full KV state for Qwen3-8B. Source positions can be compressed further by sending their token IDs (\textbf{token-ID encoding}), while latent steps remain continuous vectors because they have no token IDs. The receiver prefills the token IDs, appends the generated KV positions, and then continues the task. Across 50 tasks, the token-ID encoding reduces payload by a further \(75\times\) relative to continuous rows. While this is considerable over wide-area networks (e.g., \(664\to9\,\mathrm{ms}\) on a 300\,Mbit/s Wi-Fi link), the \(55\to0.7\,\mu\mathrm{s}\) difference on the 450\,GB/s NVLink used in our evaluation is dwarfed by other costs (\Cref{fig:fanoutqa-latency-breakdown}). We therefore use continuous rows for all reported experiments; Appendix~\ref{app:transmitted-rows} gives the full payload and transfer comparison.

\subsection{Selection Behavior}
\label{sec:selection-behavior}

Before the end-to-end evaluation, we use a small held-out set of 13 FanOutQA tasks as a diagnostic to check whether the receiver's query helps identify the evidence the receiver needs, and whether that depends on model size.  Each task contains around 50K tokens of context, and we vary Qwen 3 sizes of 1.7B, 4B, 8B and 32B. We compare selectors that are receiver-independent\footnote{We adapt StreamingLLM \citep{xiao2024streamingllm}, H$_2$O \citep{zhang2023h2o}, ChunkKV \citep{liu2025chunkkv}, and KVzip \citep{kim2025kvzip} for message selection.} with the two receiver-conditioned selectors, mean query attention and CacheBack. For each selector and position budget, we use the task's annotated evidence to measure evidence recall in the selected positions. 
Appendix~\ref{app:selector-development-details} reports the baseline adaptations and preliminary settings.

    \begin{figure}[H]
    \centering
    \includegraphics[width=\linewidth]{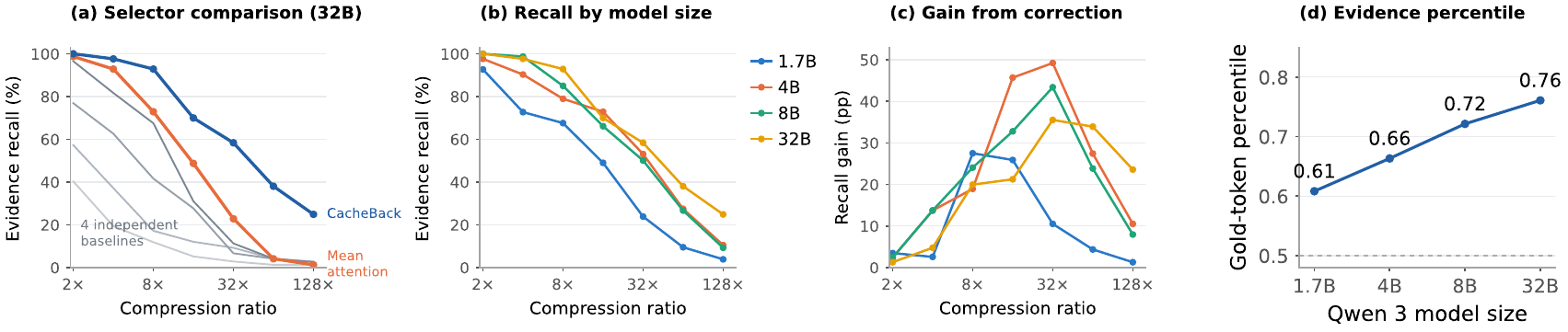}
    \caption{Results from diagnostic tests. (a) CacheBack retains high evidence recall even at $16\times$ compression compared to mean query attention and receiver-independent baselines. (b) CacheBack generally retains more evidence as model size increases. (c) The largest gains from CacheBack over mean query attention shift toward higher compression ratios with model size. (d) Mean query attention ranks gold tokens higher as model size increases.}
    \label{fig:diagostics}
\end{figure}

\paragraph{Varying selectors.}
Figure~\ref{fig:diagostics}(a) shows that as we reduce the number of positions by \(2\)--\(32\times\), mean query attention retains more evidence than every receiver-independent selector. At \(8\times\), the strongest receiver-independent selector retains \(68\%\) of the evidence, compared with \(93\%\) for CacheBack. At \(128\times\), CacheBack achieves \(25\%\) evidence recall, compared with 1--3\% for the other selectors.  We thus study CacheBack selection in our end-to-end experiments.

\paragraph{Model scale.}
\label{sec:cacheback-scale}

\Cref{fig:diagostics}(d) shows that mean query attention from the receiver query ranks gold tokens higher as model size increases. From Qwen 3 1.7B to 32B, the median gold-token percentile rises from $0.61$ to $0.76$, with $0.75$ corresponding to the top 25\% of positions.

CacheBack evidence recall also tends to improve with model scale. At \(8\times\) compression, recall rises from \(68\%\) for 1.7B to \(93\%\) for 32B. \Cref{fig:diagostics}(c) shows another effect of scale, with CacheBack’s largest gain over mean query attention shifting toward higher compression ratios in larger models.

These diagnostics show that the receiver query provides a useful signal for deciding what to send. We next test whether receiver-conditioned latent communication can improve both answer quality and latency over text communication across models and topologies.

\section{End-to-End Evaluation}
\label{sec:end-to-end-evaluation}  

We test whether latent communication improves answer quality and latency over text when both use the receiver query. Text workers use the query to generate a report, while CacheBack uses it to select sender state. The comparison therefore changes how the message is constructed, not the information need used to construct it. Where the receiver context window permits it, we also evaluate a full rows control that sends every position without selection, using CacheBack's representation. At \(16\times\) compression, CacheBack removes \(94\%\) of sender positions and has lower latency and higher accuracy than same text agents across every tested model family and communication topology (benchmark).

\subsection{Experimental Setup}

Our FanOutQA setting \citep{zhu-etal-2024-fanoutqa} tests a parallel fan-in: three agents read different Wikipedia pages, then send messages to one receiver to answer the question. For LongBench v2 \citep{bai-etal-2025-longbench}, we construct a sequential chain in which each agent reads the next part of a long document with the preceding agent’s message.

We evaluate Qwen 3 (1.7B/4B/8B) \citep{yang2025qwen3}, Ministral 3 (3B/8B/14B) \citep{liu2026ministral3}, Gemma 4 (E2B/E4B/12B) \citep{gemmateam2026gemma4}, and Nemotron Nano 2 (9B/12B) and 3 Nano (4B) \citep{nvidia2025nemotronnano2,nvidia2026nemotron3nano4b}. FanOutQA uses all four families, while LongBench v2 uses Qwen and Nemotron, covering a dense Transformer and a hybrid architecture. Within each family, the receiver is fixed to the largest model. CacheBack also uses the largest model as sender and varies from $2\times$ to $128\times$ compression on FanOutQA and from $4\times$ to $128\times$ on LongBench v2. Text baselines use either the same model, which we call \textbf{same-size text}, or a smaller sender. We compare only within each family. For hybrid architectures, CacheBack scores positions using only their global-attention layers.

Every configuration runs on an $8\times$ 80\,GB H100 GPU node. We choose the sender--receiver GPU allocation separately for each communication channel and keep that allocation fixed within the configuration. Models use their native chat templates, thinking modes, and recommended decoding settings. Text message lengths are not set from the latent message budgets.

For each benchmark, we evaluate on 50 tasks that are submitted as a batch. We report p50 and p95 time to end of answer (TTEOA), including queueing, sender computation, message preparation and transfer, receiver prefill, and receiver generation. Appendix~\ref{app:serving-settings} reports the serving settings, and Appendix~\ref{app:prompts} gives the prompts and input construction.

\begin{figure}[h]
\centering
\includegraphics[width=\linewidth]{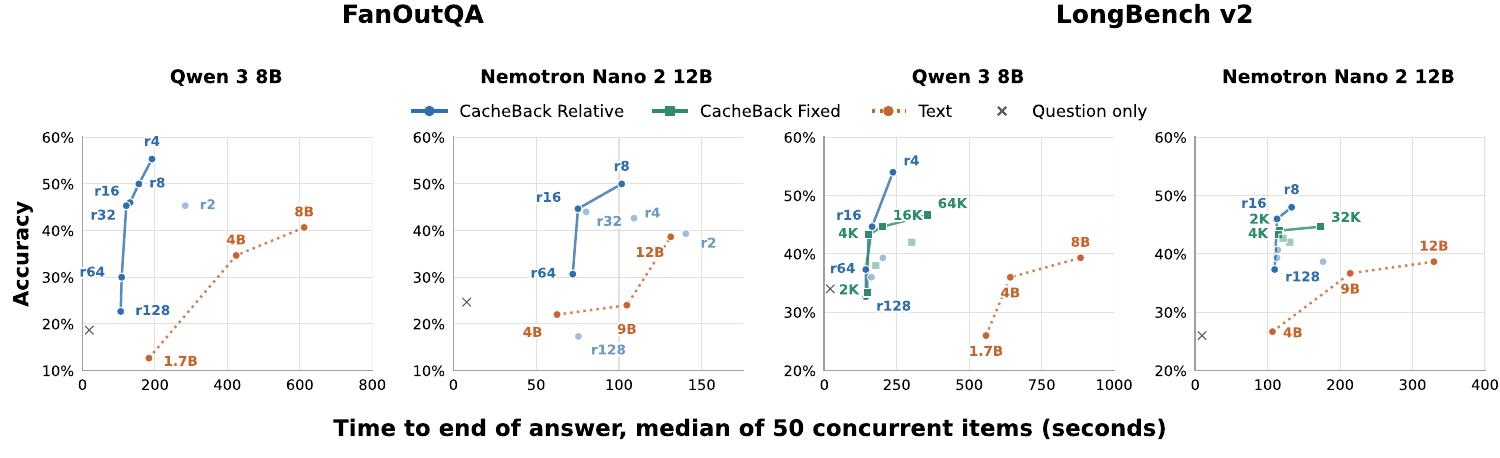}
\caption{CacheBack improves accuracy--latency tradeoffs on FanOutQA (strict accuracy) and LongBench v2 (answer accuracy). $rX$: $X$-fold compression; $x$K: fixed budget of $x$ thousand positions. Faded points are off-channel frontiers; LongBench labels mark frontier points only.}
\label{fig:combined-frontiers}
\end{figure}

Figure~\ref{fig:combined-frontiers} summarizes the Qwen and Nemotron accuracy--latency tradeoffs on both benchmarks. Appendix~\ref{app:fanoutqa-results} shows the remaining FanOutQA curves and Appendix~\ref{app:longbench-draws} the full LongBench results.

\subsection{FanOutQA: Parallel Communication}
\label{sec:fanoutqa}

This experiment tests a fan-in topology where one receiver aggregates messages from three senders. We sample 50 FanOutQA tasks from the official dev split, excluding the questions used for selector development. We keep only questions whose pages provide enough text for the split below, then take the first 50 by ascending SHA-256 hash of the question ID so no task is hand-picked. For each task, we partition the source pages among three senders, with at least 40K source tokens per sender and 120,000 in total. The receiver answers from their messages. We report strict accuracy, which requires every reference-answer group to appear in the answer. Appendix~\ref{app:answer-scoring} gives the question-selection and scoring details.

Table~\ref{tab:selected-operating-points} shows that CacheBack improves strict accuracy by 7.3--20.7 percentage points and reduces median TTEOA by
\(1.3\times\)--\(8.0\times\) relative to same-size text. For Qwen and Ministral, CacheBack $32\times$ outperforms all tested text settings on both latency and accuracy. For Gemma and Nemotron, smaller text models are faster but sacrifice considerable accuracy. Figure~\ref{fig:combined-frontiers} shows the Qwen and Nemotron curves; Figure~\ref{fig:fanoutqa-frontiers} in Appendix~\ref{app:fanoutqa-results} includes all four families.

\begin{table}[h]
\centering
\caption{Selected FanOutQA operating points versus same-size text. CB: CacheBack.}

\label{tab:selected-operating-points}

\setlength{\tabcolsep}{2pt}
\renewcommand{\arraystretch}{1}
\begin{tabular*}{\linewidth}{@{\extracolsep{\fill}}lrrrrrrrrr@{}}
& \multicolumn{3}{c}{\cellcolor{whitesmoke}Strict accuracy (\%)}
& \multicolumn{3}{c}{\cellcolor{whitesmoke}Median TTEOA (s)}
& \multicolumn{3}{c}{\cellcolor{whitesmoke}Message recall (\%)} \\
Model (ratio) & Text & CB & Gain (pp) & Text & CB & Speedup & Text & CB & Gain (pp) \\
\midrule
Qwen 3 8B ($4\times$)
& 40.7 & 55.3 & \cellcolor{gainFill}\textbf{+14.7}
& 612 & 192 & \cellcolor{gainFill}$\mathbf{3.2\times}$
& 89.5 & 95.8 & \cellcolor{gainFill}\textbf{+6.3} \\
Ministral 3 14B ($32\times$)
& 38.7 & 59.3 & \cellcolor{gainFill}\textbf{+20.7}
& 1,537 & 193 & \cellcolor{gainFill}$\mathbf{8.0\times}$
& 87.5 & 91.5 & \cellcolor{gainFill}\textbf{+4.0} \\
Gemma 4 12B ($4\times$)
& 52.0 & 59.3 & \cellcolor{gainFill}\textbf{+7.3}
& 603 & 201 & \cellcolor{gainFill}$\mathbf{3.0\times}$
& 91.1 & 95.8 & \cellcolor{gainFill}\textbf{+4.7} \\
Nemotron Nano 2 12B ($8\times$)
& 38.7 & 50.0 & \cellcolor{gainFill}\textbf{+11.3}
& 131 & 102 & \cellcolor{gainFill}$\mathbf{1.3\times}$
& 84.8 & 95.6 & \cellcolor{gainFill}\textbf{+10.8} \\
\end{tabular*}\par\vspace{3pt}
\end{table}

\begin{figure}[h]
\centering
\includegraphics[width=\linewidth]{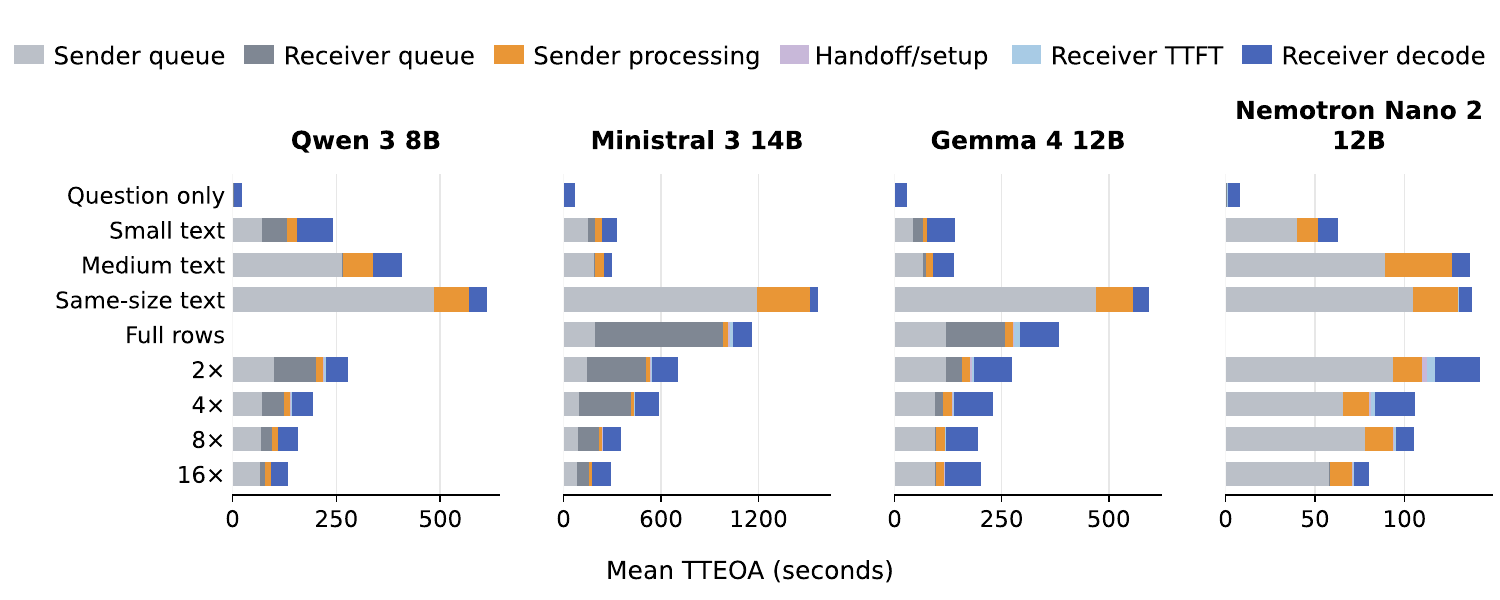}\par\nointerlineskip

\caption{FanOutQA mean TTEOA by stage: queueing, agent processing, message preparation and transfer, receiver prefill, and generation. Breakdowns beyond \(16\times\) are omitted as they change little.}
\label{fig:fanoutqa-latency-breakdown}
\end{figure}

Text messages are compact, but each additional token requires another autoregressive decoding step. CacheBack selects positions from state the agent has already computed, so it can transmit representations of more positions without generating them one at a time. At the selected operating points, these larger messages remain far smaller than the complete state and achieve \(91.5\%\)--\(95.8\%\) message recall, compared with \(84.8\%\)--\(91.1\%\) for same-size text (Table~\ref{tab:selected-operating-points}). The additional positions increase transfer and receiver prefill, but text spends more time generating its message and incurs more sender queueing under the concurrent workload (Figure~\ref{fig:fanoutqa-latency-breakdown}).

\par\addvspace{\baselineskip}
\noindent\begin{minipage}{\linewidth}
\begin{minipage}[t]{0.33\linewidth}
\vspace{0pt}
We next isolate the role of the receiver query. We replace it with a fixed unrelated query, keeping sender computation, the message budget, and the final task question unchanged. Message recall falls at every compression ratio. At \(16\times\), message recall falls from \(91\%\) to \(65\%\), while strict accuracy falls from \(46\%\) to \(23\%\) (Figure~\ref{fig:query-ablation}). The receiver query therefore changes which evidence is retained and, in turn, final accuracy.
\end{minipage}\hfill
\begin{minipage}[t]{0.6\linewidth}
\vspace{0pt}
\centering
\includegraphics[width=\linewidth]{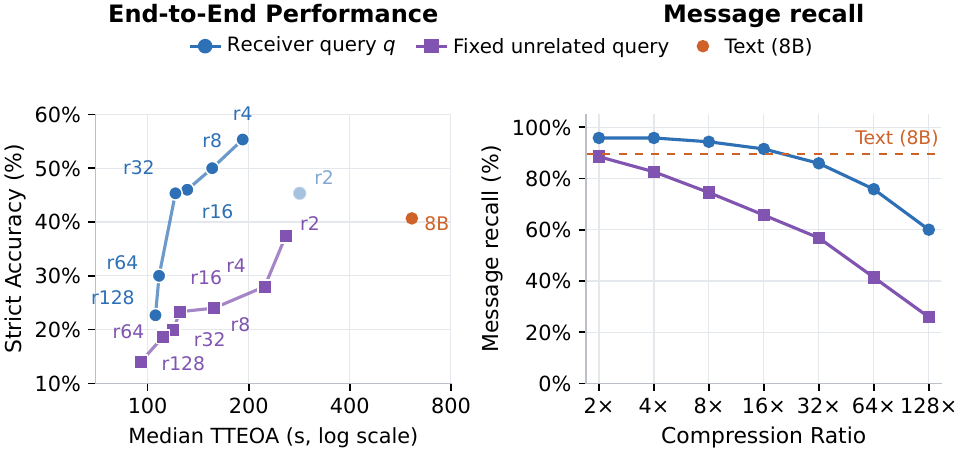}\par\nointerlineskip
\captionof{figure}{Effect of the receiver query $q$ on FanOutQA for Qwen 3 8B. The unrelated query is ``Why was Le Chaton Fat denied a small-business loan?''}
\label{fig:query-ablation}
\end{minipage}
\end{minipage}
\par\addvspace{\baselineskip}

Beyond the selected operating points, further compression yields little latency benefit and can sharply reduce accuracy (Figure~\ref{fig:combined-frontiers}, FanOutQA). For Ministral and Gemma, all sender positions fit in the receiver context, so we also send every position using the same representation as CacheBack. Comparing this with CacheBack isolates the effect of selection. Appendix~\ref{app:fanoutqa-results} reports the full results.

\subsection{LongBench v2: Sequential Communication}
\label{sec:longbench}

We construct a sequential chain to test communication when information must survive multiple agent interactions. We split each 100K–246K-token LongBench v2 document into four equal parts of 25K–61K tokens. Four agents read these parts in order. Each agent receives the previous message, reads its own part, and selects positions from the combined state to form the next message. A final receiver answers from the fourth message. Information from early parts must therefore survive several rounds of message construction before reaching the receiver.

We evaluate this chain on 50 LongBench v2 questions. In an initial run, many text sender messages hit generation limits. For simplicity, we restrict the final panel to the Easy subset as one of several changes that reduce message truncation. This preserves a fair comparison between text and latent communication. Appendix~\ref{app:longbench-evaluation} gives the full evaluation changes. Since each message is appended to the receiver's context, its size grows linearly with chain length and the context added at each step. If each agent adds \(H\) positions, the message grows to \(O(HN)\) after \(N\) steps (Section~\ref{sec:communication-costs}).

To control this growth, we evaluate two message budgets. Under the relative budget with compression ratio \(r\), the budget increases at each agent by \(1/r\) of the positions that agent adds. If each agent adds \(H\) positions, the message budget therefore grows by \(H/r\) per agent and reaches \(NH/r\) after \(N\) agents. Under the fixed budget \(B\), the outgoing message contains at most \(B\) positions, so its size stops growing once \(B\) is reached. Under both budgets, each agent reranks inherited and newly added positions together, then selects up to the budget. Inherited positions can therefore be dropped from later messages. Appendix~\ref{app:longbench-prompts} gives the prompts and input construction.

The LongBench panels in Figure~\ref{fig:combined-frontiers} show that CacheBack shifts the accuracy--latency frontier, as on FanOutQA. For Qwen, Relative at \(4\times\) and \(16\times\) achieves higher accuracy and lower latency than every text baseline. For Nemotron, a smaller text agent remains faster but less accurate. Figure~\ref{fig:longbench-frontiers} in 
Appendix~\ref{app:longbench-draws} provides the fully labeled curves.

\begin{figure}[h]
\centering
\begin{minipage}{\linewidth}
\centering
\captionof{table}{Selected operating points on LongBench v2 Easy versus same-size text. CB: CacheBack.}
\label{tab:longbench-operating-points}
\setlength{\tabcolsep}{2pt}
\renewcommand{\arraystretch}{1.0}
\begin{tabular*}{\linewidth}{@{\extracolsep{\fill}}llrrrrrr@{}}
& & \multicolumn{3}{c}{\cellcolor{whitesmoke}Accuracy (\%)} & \multicolumn{3}{c}{\cellcolor{whitesmoke}Median TTEOA (s)} \\
Model & Setting & Text & CB & Gain (pp) & Text & CB & Speedup \\
\midrule
Qwen 3 8B & Relative $4\times$
& 39.3 & 54.0 & \cellcolor{gainFill}\textbf{+14.7}
& 884 & 237 & \cellcolor{gainFill}$\mathbf{3.7\times}$ \\
& Fixed 64K
& 39.3 & 46.7 & \cellcolor{gainFill}\textbf{+7.3}
& 884 & 355 & \cellcolor{gainFill}$\mathbf{2.5\times}$ \\
Nemotron Nano 2 12B & Relative $8\times$
& 38.7 & 48.0 & \cellcolor{gainFill}\textbf{+9.3}
& 329 & 133 & \cellcolor{gainFill}$\mathbf{2.5\times}$ \\
& Fixed 32K
& 38.7 & 44.7 & \cellcolor{gainFill}\textbf{+6.0}
& 329 & 173 & \cellcolor{gainFill}$\mathbf{1.9\times}$ \\
\end{tabular*}
\end{minipage}

\end{figure}

Table~\ref{tab:longbench-operating-points} shows that \textbf{Relative} CacheBack improves accuracy by 14.7 points for Qwen and 9.3 for Nemotron, with $3.7\times$ and $2.5\times$ lower median TTEOA. \textbf{Fixed} CacheBack improves accuracy by 7.3 and 6.0 points, with $2.5\times$ and $1.9\times$ lower median TTEOA. In the sequential chain, each agent must finish its message before the next can begin. Text therefore pays autoregressive decoding at all four steps, increasing both sender processing and queueing. CacheBack avoids this decoding cost (Appendix~\ref{app:longbench-timing}).

More aggressive Relative compression yields diminishing latency benefits and degrades accuracy. For Qwen, increasing the compression ratio from $16\times$ to $128\times$ reduces median TTEOA from 165 to 143 seconds, but lowers accuracy from $44.7\%$ to $32.7\%$.

\section{Discussion and Limitations}
\label{sec:discussion}

Receiver-conditioned latent communication retains the context-management benefit of delegation without autoregressive text generation. The sender constructs its message by selecting positions from state it has already computed. While sending more selected positions increases transfer volume, receiver memory use, and prefill work, selecting them requires no additional sender decoding. The sender can therefore send more of its computed state without decoding a longer text message.

Avoiding sender decoding saves the most time when agents would otherwise spend a large share of task time decoding text messages. Long or repeated messages create this condition when they delay dependent work or keep sender GPUs busy under concurrent load. Repository exploration is one example, where several agents may need to return detailed findings before planning can continue. By contrast, an editing agent may spend most of its time changing files, running tests, or using tools before returning a short message. The advantage narrows when network transfer or receiver prefill costs more than the sender decoding avoided.

Because \(q\) is applied after sender computation, the same completed state can be scored for different receiver queries. Each new query requires another selection, but not rerunning the sender’s task or decoding a separate text message. One sender computation could therefore support several receiver-specific messages or follow-up queries from the same receiver. We do not evaluate how many such queries a sender can serve before selection and transfer become limiting.

Our primary goal is to isolate whether conditioning communication on the receiver’s information need can limit context re-expansion while preserving the benefits of latent communication. We therefore set the receiver query to the task question, issue it once after sender computation, and fix the message budget before execution. Across our long-context QA workloads, the receiver query changes which information is selected and improves the quality--latency tradeoff. This isolates the effect of receiver conditioning, but does not determine how queries should be written, when a receiver should ask again, or how much state it should request. A receiver could instead ask for evidence about an unresolved claim, request a specific file or dependency, or ask for more detail after an initial message. The query and message budget could then change as the receiver’s task changes.

We make the same choice at the level of the selector. CacheBack is deliberately simple and training-free so that the experiment tests receiver conditioning without requiring a learned communication module. It uses a fixed attention-based scoring rule, and hybrid models expose only their global-attention layers to that rule. Despite these restrictions, receiver conditioning shifts the quality--latency frontier across every tested model family and topology. These experiments do not establish the best selector. Learning the scoring rule or using more of the model state may improve selection further.

\section{Conclusion}
\label{sec:conclusion}
The emergence of latent communication has shown that changing message representation can avoid text generation and improve accuracy and latency. While these gains come from representation, representation alone does not determine what information should be sent. Full-cache latent methods still send every position in the sender sequence, so the receiver can accumulate the context that was partitioned across agents. Receiver-conditioned communication instead separates these choices. CacheBack shows that even a simple, training-free method can shift the quality--latency frontier across model families and communication topologies. A communication protocol should therefore decide not only how a message is represented, but what information it contains. And what information it contains should depend on what the receiving agent needs.

\newpage

\section*{Ethics Statement}

This work involves no human subjects or crowd workers. All experiments draw on two public benchmarks: FanOutQA, which is built from English Wikipedia, and LongBench~v2, whose documents were collected from public sources and whose questions were written by the benchmark's own annotators. We redistribute prepared FanOutQA inputs and provide scripts to reconstruct the LongBench~v2 inputs. We are aware of no fairness or bias concerns specific to this work beyond those already present in the underlying models and benchmarks. Two foreseeable risks arise from compressing communication between agents. First, a compressed handoff can discard detail that the receiver needed, which can lead to downstream errors. Second, a latent message offers no confidentiality guarantee, because it is derived from the worker's private context. Latent messages are not human-readable. We follow LatentMAS's latent protocol, whose debug mode generates parallel text probes for inspection \citep{zou2025}. The authors declare no competing interests.

\section*{Reproducibility Statement}

Appendix~\ref{app:cacheback-implementation} defines the selector and how each message is constructed, including the position budget at every compression ratio. Appendices~\ref{app:serving-settings} and~\ref{app:prompts} give the prompts issued to workers and receivers, the benchmark preparation, the settings for every reported run (model checkpoints, decoding parameters, chunking, and seeds), the scoring procedure, and the serving configuration of the eight-GPU node on which latency was measured. Our evaluation code and configurations for FanOutQA and LongBench v2, prepared FanOutQA inputs, and scripts to reconstruct the LongBench v2 inputs are available in our \href{https://github.com/maxr0ssi/rclc}{GitHub repository}. Reruns may not match our numbers exactly. Sampled decoding and batch-dependent numerics in vLLM under concurrent load can change individual outputs even with fixed seeds, and latency depends on hardware, interconnect, vLLM version, and scheduling. Appendix~\ref{app:fanoutqa-sampling} reports receiver sampling variation across three draws.

\section*{AI Use Statement}
The research idea, method, and experimental design are the authors' own; this is not automatically generated research. AI assistants were used, under the authors' direction, for implementation, running experiments, preparing figures and tables, surveying related work, and editing the manuscript. All code and pipelines were reviewed by the authors, every reported number traces to a result artifact the authors inspected, and the authors take full responsibility for the final work.

\section*{Acknowledgements}

This research received funding from NSF 2103794, 2312991, 2551201 as well as DAPLab corporate support in the form of funding and/or compute from Amazon, IntellectAI, Infosys, Tidalwave, Veris, Shopify, Microsoft, Thinking Machines, Dandy, Perplexity, and Daytona. The views and conclusions presented here are those of the authors and should not be interpreted as representing the official positions of the funding organizations.
\newpage
\bibliographystyle{iclr2027_conference}
\bibliography{references}
\newpage
\appendix

\section*{Appendix: Table of Contents}
\begingroup
\gdef\appPreviousPage{}
\newcommand{\appPage}[1]{%
  \edef\appCurrentPage{\getpagerefnumber{#1}}%
  \ifx\appCurrentPage\appPreviousPage\else
    \pageref*{#1}%
    \xdef\appPreviousPage{\appCurrentPage}%
  \fi}

\setlength{\parindent}{0pt}
\newcommand{\appentry}[1]{\gdef\appPreviousPage{}\par\addvspace{7pt}\noindent\hyperref[#1]{\textbf{\ref*{#1}\quad\nameref*{#1}}\nobreak\hfill\textbf{\appPage{#1}}}\par}
\newcommand{\appsubentry}[1]{\noindent\hspace*{1.5em}\hyperref[#1]{\ref*{#1}\quad\nameref*{#1}\nobreak\hfill\appPage{#1}}\par}
\appentry{app:communication-model}\appentry{app:obf-scope}
\appentry{app:cacheback-implementation}
{\appsubentry{app:general-scoring}}
{\appsubentry{app:budget-selection}}
{\appsubentry{app:transmitted-rows}}
\appentry{app:selector-development-details}
\appsubentry{app:evidence-retention}
\appsubentry{app:parameter-sensitivity}
\appentry{app:fanoutqa-sampling}
\appsubentry{app:fanoutqa-message-recall}
{\appsubentry{app:fanoutqa-results}}
\appsubentry{app:fanoutqa-draws}
\appsubentry{app:longbench-draws}
{\appsubentry{app:longbench-timing}}
\appentry{app:serving-settings}
\appsubentry{app:timing-output}
\appsubentry{app:answer-scoring}
\appsubentry{app:longbench-evaluation}
\appentry{app:prompts}
\appsubentry{app:fanoutqa-prompts}
\appsubentry{app:longbench-prompts}
\appsubentry{app:chat-packing}
\endgroup
\clearpage

\section{Formal Model of Receiver-Conditioned Communication}
\label{sec:receiver-conditioned-model}\label{app:communication-model}

We formalize the communication choices introduced in
Section~\ref{sec:communication-model}. A receiving agent provides a query \(q\) describing the information it seeks from a completed agent state. Message construction then separates two choices: what information is selected and how that information is represented for transmission.

\paragraph{Communication model.}
A sender \(S_\theta\) and receiver \(R_\theta\) share model weights 
\(\theta\) (or a mapping between their representations) and cooperate on task \(T\). The sender processes input
\(x\), producing a layer-wise KV cache, where \(j\) indexes
positions in the sender sequence:
\begin{equation}
    \mathcal M_S
    =
    \left\{
        (K^{(\ell)}_j,V^{(\ell)}_j)
    \right\}_{\ell=1,\ldots,L;\;j=1,\ldots,N}
\end{equation}

Let \(r\) denote the receiver's current context. The content map
\(C_\phi\) determines what information to communicate for the
query \(q\). The representation map \(P_\psi\) determines its
transmitted form:
\begin{equation}
    c=C_\phi(\mathcal M_S;q),
    \qquad
    \tau=P_\psi(c),
\end{equation}
The receiver decodes the message and uses it to produce its output,
\begin{equation}
    \hat y=R_\theta(T,r;D_\psi(\tau)).
\end{equation}

\paragraph{Quality and cost.}
For task target \(y\) and loss \(\ell\), we measure message adequacy
by the receiver's expected task loss,
\begin{equation}
    \mathcal E=\mathbb E[\ell(\hat y,y)],
\end{equation}
averaging over tasks and generation randomness. Let \(J\) be a
measured cost under fixed models, hardware, and workload, such as
median task-completion latency. This latency includes worker
computation, message construction and transfer, receiver processing,
and scheduling delays. Both cost and task loss depend on the content
map, representation map, and position budgets
\(\mathbf B=(B_1,\ldots,B_M)\). Among compatible implementations
satisfying Equation~\ref{eq:receiver-capacity}, we seek
\begin{equation}
    \underbrace{\min_{\phi,\psi,\mathbf B} J}
        _{\text{minimize cost}}
    \qquad \text{subject to} \qquad
    \underbrace{
        \vphantom{\min_{\phi,\psi,\mathbf B}}
        \mathcal E\leq\delta
    }_{\text{meet the quality target}},
    \label{eq:communication-objective}
\end{equation}
where \(\delta\) is the maximum acceptable task loss. Varying
\(\delta\) traces the quality--cost frontier for the chosen models,
workload, and design space.

A text report chooses content through autoregressive generation. When generated for the receiver's query \(q\), it implements a receiver-conditioned content map. A report generated only for the worker's original assignment may omit information needed for a different receiver query. Receiver conditioning is therefore a property of message construction, not of whether the message is textual or latent.

LatentMAS instead fixes
\(C_\phi(\mathcal M_S;q)=\mathcal M_S\) \citep{zou2025}.
It returns the entire state rather than choosing information for
the receiver query. Layer selection and input-embedding transfer
can reduce the transmitted payload while retaining all source
positions. When \(B_i<T_i\), however, message construction must
also choose which positions to omit. A choice made without \(q\)
cannot adapt to different queries over the same worker state.

\section{Comparison with OBF}
\label{app:obf-scope}
OBF studies compressed KV relay in a sequential reasoning chain whose agents receive the same question~\citep{li2026obf}. Our experiments instead distribute source evidence across workers: different articles in FanOutQA and successive document chunks in LongBench v2. The receiver must obtain information from those private inputs through the handoffs.
\begin{table}[h]
\centering
\definecolor{figgray}{HTML}{666A70}
\caption{Information access and context growth. Shared-question reasoning versus communication of private evidence.}
\label{tab:obf-scope}
\vspace{6pt}
\normalsize
\renewcommand{\arraystretch}{1.13}
\setlength{\tabcolsep}{5pt}
\begin{tabularx}{\linewidth}{@{}p{0.17\linewidth}*{2}{>{\raggedright\arraybackslash}X}@{}}
\toprule
& \textbf{OBF evaluation} & \textbf{Our evaluation} \\
\midrule
Information
& Same question at every agent; no private source documents
& Agents receive distinct evidence unavailable to the receiver \\
\addlinespace[5pt]
Selection signal
& Attention from current latent reasoning states
& Attention from the receiver's query \\
\addlinespace[5pt]
Topology
& Sequential reasoning chain
& Sequential (LongBench v2) and parallel (FanOutQA) \\
\addlinespace[5pt]
Total input
& 496--1,012 prompt tokens across three reasoning agents
& FanOutQA: 122k--150k evidence tokens; LongBench v2: 100k--246k source tokens \\
\addlinespace[5pt]
Context growth
& Inherited state is preserved; retained prompt KV and new thoughts accumulate
& Relative budgets grow with added context; fixed budgets cap the message size. Fan-in combines compressed agent messages \\
\bottomrule
\end{tabularx}\par
\vspace{6pt}
{\raggedright\normalsize
Input counts: OBF reports mean prompt tokens, including repeated prompts; ours count source tokens once per task using the Qwen tokenizer.\par}
\end{table}
\FloatBarrier

\section{Selector and message definitions}
\label{app:cacheback-implementation}
\subsection{General scoring rule}
\label{app:general-scoring}
\label{app:cacheback-scoring}

\paragraph{Adapting SnapKV.}
We adapt SnapKV's attention-based scoring~\citep{li2024snapkv}, using the receiver's query in place of the sender's prompt suffix. The sender computes attention from the query \(q\) to its completed KV cache. Only attention from query tokens is used to compute the selection scores. Let \(m\) be the number of query tokens, \(\mathcal L\) the set of attention layers used for scoring, and \(\mathcal H_\ell\) the query heads in layer \(\ell\). Write \(a_{\ell,h,r,t}\) for attention from query token \(r\), in query head \(h\) of layer \(\ell\), to sender position \(t\). If a model uses both local and global attention, we use only global-attention layers for scoring. We use \(\operatorname{Pool}_7\) to denote a local maximum over seven sender positions, centered on position \(t\) and clipped at sequence boundaries. Pooling is only over sender positions.

\paragraph{Mean query attention.}
For each layer in \(\mathcal L\), we average attention over query tokens and query heads. We then sum across layers and pool over sender positions:
\begin{equation}
Q_t
=
\operatorname{Pool}_7\!\left(
\sum_{\ell\in\mathcal L}
\frac{1}{m|\mathcal H_\ell|}
\sum_{r=1}^{m}
\sum_{h\in\mathcal H_\ell}
a_{\ell,h,r,t}
\right).
\label{eq:mean-query-attention}
\end{equation}
This gives one shared ranking of sender positions across query heads.

\paragraph{CacheBack's correction.}
Averaging across query tokens can hide a sender position that receives strong attention from one part of the query. CacheBack compares attention before and after this averaging. We first average query heads that share each KV head. Let \(\mathcal S\) be the set of these KV-head groups across layers in \(\mathcal L\), and let \(b_{s,r,t}\) be the mean attention in KV-head group \(s\) from query token \(r\) to sender position \(t\). For \(p>1\), define
\begin{equation}
C_{p,t}=\left(
\frac{
\sum_{s\in\mathcal S}\operatorname{mean}_r
\left[\operatorname{Pool}_7(b_{s,r,t})^p\right]
}{
\sum_{s\in\mathcal S}
\left[\operatorname{Pool}_7\!\left(\operatorname{mean}_r b_{s,r,t}\right)\right]^p
}
\right)^{1/(p-1)},
\qquad
C_t\equiv C_{2,t}.
\label{eq:cacheback-correction}
\end{equation}
For the numerator, we pool over sender positions and raise to \(p\) before averaging over query tokens. For the denominator, we average over query tokens before pooling and taking the \(p\)th power. CacheBack uses \(S_t^{\mathrm{CB}}=Q_t C_t^2\). Appendix~\ref{app:parameter-sensitivity} varies \(p\) and how strongly \(C_{p,t}\) changes the score.

\paragraph{Interpretation.}
At \(p=2\), without pooling, the correction reduces to one plus a normalized variance across query tokens:
\begin{equation}
C_t=1+
\frac{\sum_s\operatorname{Var}_r(b_{s,r,t})}
{\sum_s(\operatorname{mean}_r b_{s,r,t})^2}.
\label{eq:appendix-unpooled-variance}
\end{equation}
The identity holds when the denominator is nonzero; \(\operatorname{Var}_r\) is the population variance over query tokens. With pooling, query tokens can peak at different positions within the same seven-position window, which can also increase \(C_t\). When both sums are zero, we set \(C_{p,t}=1\).

\subsection{Budgets and position selection}
\label{app:budget-selection}
\label{app:cacheback-selection}

Mean query attention and CacheBack select positions using the same rule. We keep the first position and all 40 current latent positions, then add positions from ranked spans until the budget is full. These 41 positions count toward the budget. For FanOutQA, a sender $i$ with $T_i$ prompt and latent positions selects
\begin{equation}
B_i=\left\lceil T_i/r\right\rceil
\label{eq:appendix-fanout-budget}
\end{equation}
positions at compression ratio $r$, so $B_i-41$ positions remain for span selection. The count $T_i$ includes prompt framing and the 40 latent positions. Every evaluated budget contains at least 41 positions.

In a sequential chain, let $T_i^{\mathrm{old}}$ count inherited positions and $T_i^{\mathrm{new}}$ count newly added prompt and latent positions. We evaluate two budget rules,
\begin{equation}
B_i^{\mathrm{relative}}=T_i^{\mathrm{old}}+\left\lceil T_i^{\mathrm{new}}/r\right\rceil,
\qquad
B_i^{\mathrm{fixed}}=\min(B_{\max},T_i^{\mathrm{old}}+T_i^{\mathrm{new}}).
\label{eq:appendix-chain-budgets}
\end{equation}
For $B_i^{\mathrm{relative}}$, inherited positions increase the budget, but inherited and new positions are ranked together. We always keep the first position and the 40 current latent positions. Inherited latent positions can be discarded. Discarded positions are not available in later messages.

Spans begin at position zero, have width 16, and are ranked by their mean score. The last span may be shorter. Before ranking spans, we set the first position's score to the maximum position score. Span means include the first position, so its maximum score contributes to the first span's mean score. Equal span scores favor the earlier span.

We add each span's unselected positions in rank order until the budget is full. If the next span would exceed the budget, we add a shorter interval around its highest-scoring position. We return selected positions in source order.

\subsection{Message encoding}
\label{app:transmitted-rows}
\label{app:cacheback-rows}

\begin{figure}[h]
\centering
\includegraphics[width=0.9\linewidth]{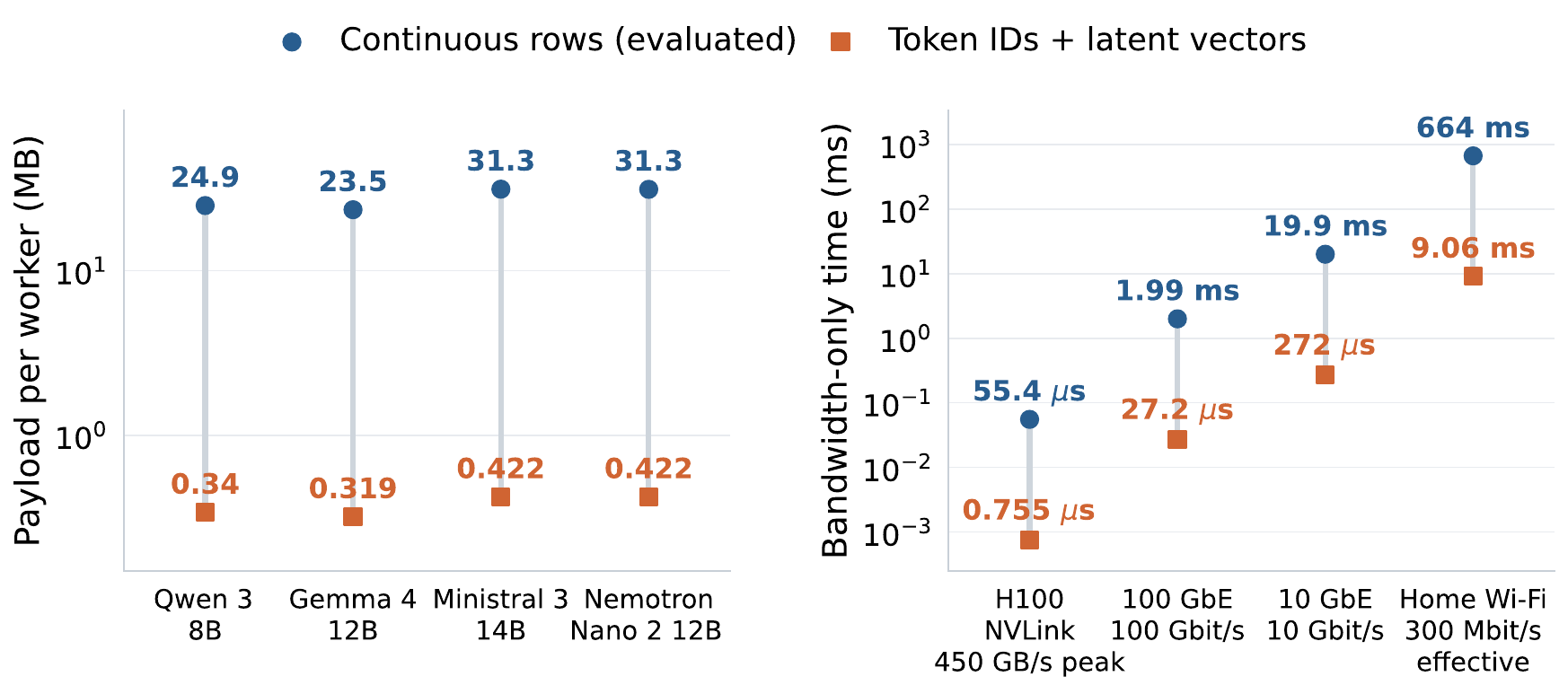}
\caption{Continuous-row versus token-ID encoding with latent vectors. Left: payload per sender. Right: bandwidth-only transfer time. Both panels use log scales.}
\label{fig:message-size}
\end{figure}

Section~\ref{sec:communication-model} separates what is sent from how it is represented. Our experiments transmit every selected position as one continuous row and reconstruct the receiver KV cache with a fresh prefill.
Because we re-prefill the selected rows, the receiver applies positional encodings at their new positions; direct KV transfer would require handling the sender--receiver positional mismatch explicitly, which we do not evaluate. The token-ID encoding instead sends selected source positions as token IDs and latent positions as continuous vectors. For the same model, the receiver reconstructs the corresponding source embeddings exactly from those token IDs. Figure~\ref{fig:message-size} compares the resulting payloads and bandwidth-only transfer times. On the node-local links we evaluate, continuous-row transfer is small relative to other costs, so all reported experiments use it. Implementations that minimize transmitted bytes should use the token-ID encoding. Appendix~\ref{app:serving-settings} specifies scaling and alignment, and Appendix~\ref{app:prompts} specifies message placement and chat framing.

\newpage
\section{Selector development}
\label{app:selector-development-details}
\subsection{Evidence recall}
\label{app:evidence-retention}

We use 13 FanOutQA development questions for which every required piece of evidence not stated in the question can be located in the context provided to the agents. For each question, an evidence group counts as retained only when the selected positions keep every token of each required piece of evidence in that group. If the same evidence appears more than once, keeping any one copy is sufficient. Evidence recall averages the retained fraction across the 13 questions.

We compare StreamingLLM, H$_2$O, ChunkKV, KVzip, mean query attention, and CacheBack. StreamingLLM ranks positions by recency. H$_2$O sums causal prompt attention for each position. ChunkKV scores positions from the final 32 prompt rows, averages over a width-5 local window, and assigns positions the mean score of their 16-position chunk. KVzip scores sender positions using attention during context reconstruction. All six selectors rank sender positions and use the same position-selection rule and budget from Appendix~\ref{app:budget-selection}. We fixed width-16 spans and \(\operatorname{Pool}_7\) during preliminary development before this comparison rather than optimizing them on these 13 questions.

\begin{figure}[h]
\centering
\includegraphics[width=\linewidth]{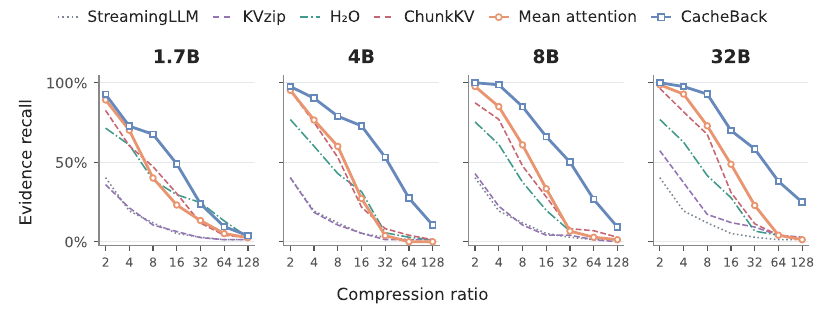}
\caption{Mean evidence recall by selector across Qwen 3 models (\(n=13\)).}
\label{fig:appendix-selector-recall}
\end{figure}

Figure~\ref{fig:appendix-selector-recall} compares evidence recall across Qwen 3 1.7B, 4B, 8B, and 32B. CacheBack has the highest recall at 26 of the 28 model-size--ratio settings. The two exceptions are Qwen 3 1.7B at \(32\times\) and \(64\times\), where H$_2$O retains more evidence. We therefore use CacheBack in the end-to-end evaluation.
\FloatBarrier

\clearpage
\subsection{Parameter sensitivity}
\label{app:parameter-sensitivity}

We vary the power $p>1$ inside the correction in Equation~\ref{eq:cacheback-correction} and the strength $\alpha\geq0$, which controls how strongly the correction changes $Q_t$. The score is
\begin{equation}
S_t(p,\alpha)=Q_t C_{p,t}^{\alpha}.
\label{eq:appendix-general-score}
\end{equation}
Mean query attention uses $\alpha=0$. CacheBack uses $p=\alpha=2$.

We selected $p=\alpha=2$ on the 13 FanOutQA development questions and kept them fixed across model sizes, compression ratios, and the end-to-end evaluation.

At $p=\infty$, the correction compares the largest pooled query-token attention with the largest value after averaging query tokens.

\begin{table}[h]
\centering
\definecolor{chosenFill}{HTML}{E8EEF6}
\caption{Evidence recall (\%) over 13 tasks and four Qwen 3 sizes. Shading marks $p=\alpha=2$; bold marks the best value at each ratio.}
\label{tab:selector-sensitivity}
\vspace{6pt}
\normalsize
\renewcommand{\arraystretch}{1.08}
\setlength{\tabcolsep}{5pt}
\begin{tabularx}{\linewidth}{>{\centering\arraybackslash}X*{8}{>{\raggedleft\arraybackslash}X}}
\toprule
$\alpha$ & $p$ & $2\times$ & $4\times$ & $8\times$ & $16\times$ & $32\times$ & $64\times$ & $128\times$ \\
\midrule
\multicolumn{9}{l}{\bfseries Mean query attention} \\
$0$ & $-$ & 95 & 81 & 58 & 33 & 12 & 3 & 1 \\
\midrule
\multicolumn{9}{l}{\bfseries CacheBack} \\
$0.25$ & $1.5$ & 96 & 86 & 67 & 41 & 15 & 3 & 1 \\
$0.5$ &  & 96 & 86 & 70 & 50 & 21 & 9 & 1 \\
$1$ &  & 95 & \textbf{90} & 78 & 57 & 33 & 16 & 4 \\
$2$ &  & \textbf{98} & \textbf{90} & \textbf{81} & \textbf{65} & \textbf{46} & 22 & 12 \\
$4$ &  & \textbf{98} & 89 & 74 & 60 & 40 & 17 & 10 \\
\addlinespace[4pt]
$0.25$ & $2$ & 96 & 85 & 67 & 41 & 15 & 3 & 1 \\
$0.5$ &  & 96 & 86 & 70 & 49 & 22 & 9 & 2 \\
$1$ &  & 94 & 88 & 76 & 57 & 34 & 16 & 5 \\
\rowcolor{chosenFill}
\textbf{2} & \textbf{2} & \textbf{98} & \textbf{90} & \textbf{81} & 64 & \textbf{46} & 25 & 12 \\
$4$ &  & 97 & 89 & 75 & 61 & 42 & 23 & 12 \\
\addlinespace[4pt]
$0.25$ & $3$ & 96 & 85 & 67 & 38 & 15 & 3 & 1 \\
$0.5$ &  & 96 & 86 & 70 & 47 & 20 & 8 & 2 \\
$1$ &  & 94 & 89 & 78 & 55 & 31 & 15 & 5 \\
$2$ &  & 96 & 89 & 78 & 64 & 42 & 25 & 12 \\
$4$ &  & 96 & 87 & 76 & 61 & 41 & 24 & \textbf{13} \\
\addlinespace[4pt]
$0.25$ & $4$ & 96 & 85 & 66 & 39 & 15 & 4 & 1 \\
$0.5$ &  & 96 & 85 & 69 & 44 & 20 & 8 & 2 \\
$1$ &  & 94 & 89 & 76 & 54 & 29 & 15 & 3 \\
$2$ &  & 96 & 89 & 77 & 63 & 42 & 25 & 12 \\
$4$ &  & 96 & 87 & 74 & 62 & 39 & 24 & 12 \\
\addlinespace[4pt]
$0.25$ & $8$ & 96 & 84 & 66 & 38 & 15 & 4 & 1 \\
$0.5$ &  & 96 & 85 & 68 & 44 & 17 & 8 & 2 \\
$1$ &  & 94 & 88 & 74 & 51 & 26 & 14 & 3 \\
$2$ &  & 96 & \textbf{90} & 76 & 63 & 41 & 20 & 11 \\
$4$ &  & 95 & 89 & 76 & 61 & 42 & 24 & 12 \\
\addlinespace[4pt]
$1$ & $\infty$ & 94 & 88 & 71 & 51 & 25 & 13 & 3 \\
$2$ &  & 96 & 89 & 77 & 61 & 40 & 20 & 9 \\
$4$ &  & 95 & 89 & 77 & 61 & 41 & \textbf{28} & 11 \\
$8$ &  & 96 & 86 & 70 & 59 & 40 & 26 & 11 \\
\bottomrule
\end{tabularx}
\end{table}

\FloatBarrier

\clearpage
\section{Results and receiver sampling}
\label{app:fanoutqa-sampling}

\subsection{FanOutQA message recall}
\label{app:fanoutqa-message-recall}

\begin{figure}[h]
\centering
\includegraphics[width=0.9\linewidth]{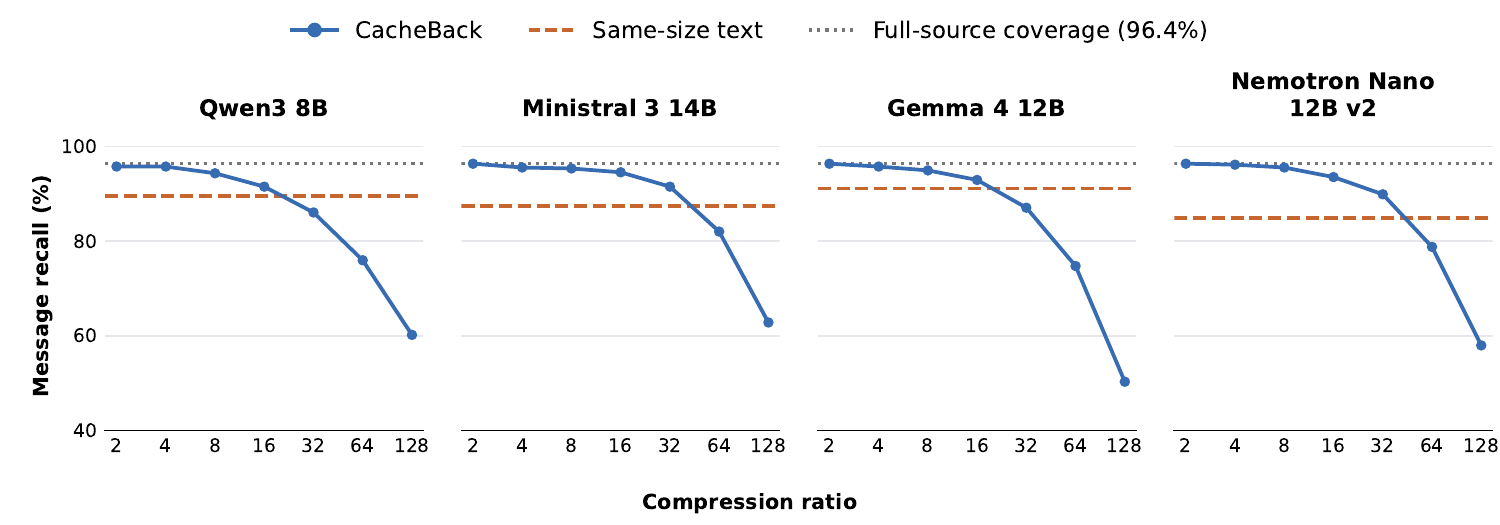}
\caption{FanOutQA message recall over 50 questions and 495 reference groups.}
\label{fig:fanoutqa-message-recall}
\end{figure}

Our goal is to measure how much reference evidence reaches the receiver in the sender messages. We therefore compute message recall before receiver generation as the fraction of the 495 FanOutQA reference groups present in the three sender messages. For text, we match accepted answer strings in the generated messages. For CacheBack, we match accepted answer strings within retained source-token spans without joining across discarded positions, and exclude generated continuous inputs. Appendix~\ref{app:answer-scoring} gives the exact matching rules. Table~\ref{tab:fanoutqa-message-recall} and Figure~\ref{fig:fanoutqa-message-recall} report recall for every setting.

For CacheBack, this matching rule has a 96.4\% ceiling. Eight reference groups appear in the question text and are included in the transmitted prompt framing. Another 18 do not appear in any sender context, so retained source tokens cannot contain them. The remaining 469 groups can be located in the sender context. Thus 477 of 495 groups can be matched. The full-row controls for Ministral and Gemma recover all 477 groups.

At $2\times$, CacheBack messages contain 474 to 477 of the 495 reference groups across the four model families. Recall remains above 91\% through $16\times$ in every family. It falls to 86.1--91.5\% at $32\times$, 74.7--82.0\% at $64\times$, and 50.3--62.8\% at $128\times$.

Same-size text messages contain 84.8--91.1\% of the reference groups across model families. At the CacheBack operating points in Table~\ref{tab:selected-operating-points}, the messages contain 91.5--95.8\%. We pool message recall over reference groups, whereas strict accuracy requires every reference group for a question. A single missing group can therefore make the entire question strictly incorrect.

\begin{table}[h]
\centering
\caption{FanOutQA message recall over 50 questions. CacheBack recall has a 96.4\% ceiling.}
\label{tab:fanoutqa-message-recall}
\vspace{5pt}
\normalsize
\setlength{\tabcolsep}{6pt}
\begin{tabular}{lrrrr}
\toprule
\textbf{Setting} & \textbf{Qwen 3} & \textbf{Ministral 3} & \textbf{Gemma 4} & \textbf{Nemotron} \\
\midrule
\multicolumn{5}{@{}l}{\textbf{Text channels}} \\
Small  & 72.1 & 82.6 & 86.7 & 74.5 \\
Medium & 86.7 & 90.5 & 87.1 & 77.6 \\
Same-size  & 89.5 & 87.5 & 91.1 & 84.8 \\
\midrule
\multicolumn{5}{@{}l}{\textbf{Latent channels (CacheBack)}} \\
Full rows         & --   & 96.4 & 96.4 & --   \\
$2\times$    & 95.8 & 96.4 & 96.4 & 96.4 \\
$4\times$    & 95.8 & 95.6 & 95.8 & 96.2 \\
$8\times$    & 94.3 & 95.4 & 94.9 & 95.6 \\
$16\times$   & 91.5 & 94.5 & 92.9 & 93.5 \\
$32\times$   & 86.1 & 91.5 & 87.1 & 89.9 \\
$64\times$   & 76.0 & 82.0 & 74.7 & 78.8 \\
$128\times$  & 60.2 & 62.8 & 50.3 & 58.0 \\
\bottomrule
\end{tabular}
\end{table}

\FloatBarrier

\subsection{Full FanOutQA results}
\label{app:fanoutqa-results}

Tables~\ref{tab:fanoutqa-full-results-global} and \ref{tab:fanoutqa-full-results-hybrid} report loose and strict accuracy,
p50 and p95 TTEOA, and speedup over same-size text for every FanOutQA setting. TTEOA runs from question submission to the end of the first
receiver answer. It includes queueing, sender computation, message
preparation and transfer, receiver prefill, and receiver decoding.
All 50 questions are submitted together, so TTEOA also includes queueing
caused by concurrent questions on the same node.

\begin{figure}[h]
\centering
\includegraphics[
    trim=0 22bp 0 0,
    clip,
    width=\linewidth
]{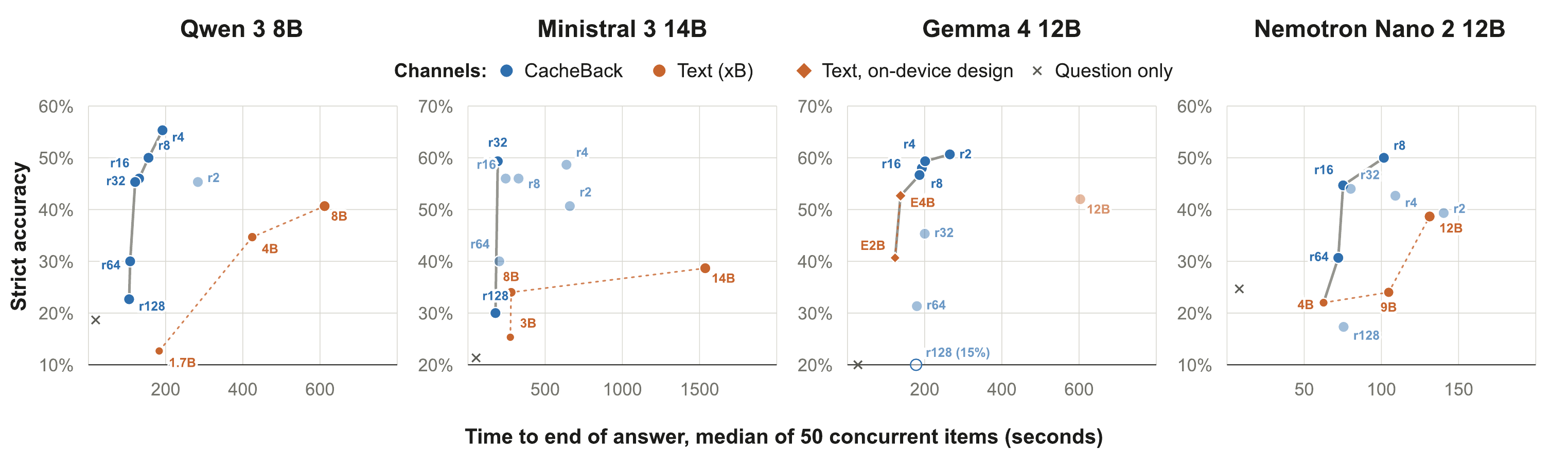}
\caption{FanOutQA strict accuracy and median TTEOA for all evaluated settings. $rX$ denotes CacheBack compression ratio $X$.}
\label{fig:fanoutqa-frontiers}
\end{figure}

Figure~\ref{fig:fanoutqa-frontiers} shows strict accuracy against median
TTEOA for all four model families.

\definecolor{fanoutTextBand}{HTML}{F8ECE6}
\definecolor{fanoutLatentBand}{HTML}{E6EDF5}

\begin{table}[h]
\centering
\caption{Full FanOutQA results for global-attention Transformers. Loose and strict accuracy (\%) over 50 questions and three receiver draws per setting.}
\label{tab:fanoutqa-full-results-global}
\vspace{5pt}
\normalsize
\setlength{\tabcolsep}{2pt}
\renewcommand{\arraystretch}{1.14}

\begin{tabular*}{\linewidth}{@{\extracolsep{\fill}}l*{10}{r}@{}}
\toprule
& \multicolumn{5}{c}{\shortstack{\textbf{Qwen 3}\\[-0.15em]8B receiver}}
& \multicolumn{5}{c}{\shortstack{\textbf{Ministral 3}\\[-0.15em]14B receiver}} \\
\cmidrule(lr){2-6}\cmidrule(lr){7-11}
& \multicolumn{2}{c}{\textbf{Accuracy} $\uparrow$}
& \multicolumn{3}{c}{\textbf{TTEOA}}
& \multicolumn{2}{c}{\textbf{Accuracy} $\uparrow$}
& \multicolumn{3}{c}{\textbf{TTEOA}} \\
\textbf{Setting}
& Loose & Strict & p50 (s) & p95 (s) & Speedup
& Loose & Strict & p50 (s) & p95 (s) & Speedup \\
\midrule

Question only
& 57 & 19 & 19 & 62 & --
& 53 & 21 & 53 & 309 & -- \\

\multicolumn{6}{@{}l}{\cellcolor{fanoutTextBand}\strut\textbf{Text channels}}
& \multicolumn{5}{r@{}}{\cellcolor{fanoutTextBand}\strut} \\

Small
& 53 & 13 & 183 & 515 & 3.3$\times$
& 63 & 25 & 274 & 709 & 5.6$\times$ \\

Medium
& 73 & 35 & 424 & 661 & 1.4$\times$
& 69 & 34 & 280 & 552 & 5.5$\times$ \\

Same-size
& 78 & 41 & 612 & 1059 & 1.0$\times$
& 74 & 39 & 1537 & 2750 & 1.0$\times$ \\

\multicolumn{6}{@{}l}{\cellcolor{fanoutLatentBand}\strut\textbf{Latent channels}}
& \multicolumn{5}{r@{}}{\cellcolor{fanoutLatentBand}\strut} \\

Full rows
& \multicolumn{5}{c}{Exceeds context limit}
& 72 & 44 & 1041 & 2296 & 1.5$\times$ \\

$2\times$
& 83 & 45 & 284 & 503 & 2.2$\times$
& 74 & 51 & 660 & 1278 & 2.3$\times$ \\

$4\times$
& \textbf{84} & \textbf{55} & 192 & 322 & 3.2$\times$
& 83 & \textbf{59} & 638 & 1048 & 2.4$\times$ \\

$8\times$
& \textbf{84} & 50 & 156 & 264 & 3.9$\times$
& 80 & 56 & 328 & 634 & 4.7$\times$ \\

$16\times$
& 81 & 46 & 131 & 221 & 4.7$\times$
& 81 & 56 & 244 & 669 & 6.3$\times$ \\

$32\times$
& 82 & 45 & 121 & 201 & 5.1$\times$
& \textbf{84} & \textbf{59} & 193 & 483 & 8.0$\times$ \\

$64\times$
& 76 & 30 & 108 & 186 & 5.7$\times$
& 80 & 40 & 203 & \textbf{471} & 7.6$\times$ \\

$128\times$
& 70 & 23 & \textbf{106} & \textbf{182} & \textbf{5.8}$\times$
& 74 & 30 & \textbf{178} & \textbf{471} & \textbf{8.6}$\times$ \\

\bottomrule
\end{tabular*}

\par\vspace{5pt}
{\raggedright\normalsize
Text sender sizes are 1.7B / 4B / 8B for Qwen 3 and 3B / 8B / 14B for Ministral 3.
Latent senders match the receiver.
Speedup is relative to same-size text.
\par}
\end{table}

\begin{table}[h]
\centering
\caption{Full FanOutQA results for hybrid architectures. Loose and strict accuracy (\%) over 50 questions and three receiver draws per setting.}
\label{tab:fanoutqa-full-results-hybrid}
\vspace{5pt}
\normalsize
\setlength{\tabcolsep}{2pt}
\renewcommand{\arraystretch}{1.14}

\begin{tabular*}{\linewidth}{@{\extracolsep{\fill}}l*{10}{r}@{}}
\toprule
& \multicolumn{5}{c}{\shortstack{\textbf{Gemma 4}\\[-0.15em]Global + sliding-window attention\\[-0.15em]12B receiver}}
& \multicolumn{5}{c}{\shortstack{\textbf{Nemotron}\\[-0.15em]Mamba + attention\\[-0.15em]12B receiver}} \\
\cmidrule(lr){2-6}\cmidrule(lr){7-11}
& \multicolumn{2}{c}{\textbf{Accuracy} $\uparrow$}
& \multicolumn{3}{c}{\textbf{TTEOA}}
& \multicolumn{2}{c}{\textbf{Accuracy} $\uparrow$}
& \multicolumn{3}{c}{\textbf{TTEOA}} \\
\textbf{Setting}
& Loose & Strict & p50 (s) & p95 (s) & Speedup
& Loose & Strict & p50 (s) & p95 (s) & Speedup \\
\midrule

Question only
& 3 & 1 & 27 & 70 & --
& 60 & 25 & 8 & 13 & -- \\

\multicolumn{6}{@{}l}{\cellcolor{fanoutTextBand}\strut\textbf{Text channels}}
& \multicolumn{5}{r@{}}{\cellcolor{fanoutTextBand}\strut} \\

Small
& 75 & 41 & \textbf{123} & 272 & \textbf{4.9}$\times$
& 59 & 22 & \textbf{63} & \textbf{110} & \textbf{2.1}$\times$ \\

Medium
& 80 & 53 & 138 & \textbf{228} & 4.4$\times$
& 66 & 24 & 105 & 331 & 1.3$\times$ \\

Same-size
& 83 & 52 & 603 & 1021 & 1.0$\times$
& 73 & 39 & 131 & 251 & 1.0$\times$ \\

\multicolumn{6}{@{}l}{\cellcolor{fanoutLatentBand}\strut\textbf{Latent channels}}
& \multicolumn{5}{r@{}}{\cellcolor{fanoutLatentBand}\strut} \\

Full rows
& \textbf{88} & \textbf{62} & 363 & 775 & 1.7$\times$
& \multicolumn{5}{c}{Exceeds context limit} \\

$2\times$
& 86 & 61 & 265 & 484 & 2.3$\times$
& 75 & 39 & 140 & 238 & 0.9$\times$ \\

$4\times$
& \textbf{88} & 59 & 201 & 412 & 3.0$\times$
& \textbf{79} & 43 & 109 & 172 & 1.2$\times$ \\

$8\times$
& 85 & 57 & 187 & 364 & 3.2$\times$
& \textbf{79} & \textbf{50} & 102 & 166 & 1.3$\times$ \\

$16\times$
& 87 & 58 & 193 & 353 & 3.1$\times$
& 77 & 45 & 75 & 139 & 1.7$\times$ \\

$32\times$
& 83 & 45 & 200 & 321 & 3.0$\times$
& \textbf{79} & 44 & 80 & 133 & 1.6$\times$ \\

$64\times$
& 71 & 31 & 180 & 302 & 3.4$\times$
& 73 & 31 & 72 & 133 & 1.8$\times$ \\

$128\times$
& 56 & 15 & 178 & 322 & 3.4$\times$
& 57 & 17 & 76 & 129 & 1.7$\times$ \\

\bottomrule
\end{tabular*}

\par\vspace{5pt}
{\raggedright\normalsize
Text sender sizes are E2B / E4B / 12B for Gemma 4 and 4B / 9B / 12B for Nemotron.
Latent senders match the receiver.
Speedup is relative to same-size text.
\par}
\end{table}
\FloatBarrier

\subsection{FanOutQA receiver sampling}
\label{app:fanoutqa-draws}

CacheBack selection is deterministic once the completed sender state and receiver query \(q\) are fixed. We therefore hold the completed sender state and selected positions fixed across the three receiver draws and vary only the receiver seed. Text sender messages are likewise fixed across draws. Strict scoring uses deterministic reference matching, so differences across draws reflect receiver generation. We report each draw and the standard deviation across the three draw-level means.

To estimate uncertainty across questions and receiver generations, we use a hierarchical bootstrap with 10,000 samples. Each sample resamples 50 questions and then three receiver draws within each sampled question. We report the 2.5th and 97.5th percentiles as the 95\% confidence interval. For CacheBack--text differences, we use the same sampled questions for both settings and resample receiver draws separately within each setting.

\begin{table}[h]
\centering
\caption{Receiver sampling on FanOutQA. Strict accuracy (\%) over 50 questions and three receiver draws per setting.}
\label{tab:fanoutqa-sampling}
\vspace{5pt}
\normalsize
\setlength{\tabcolsep}{5pt}
\renewcommand{\arraystretch}{1.08}
\begin{tabular*}{\linewidth}{@{\extracolsep{\fill}}lrrrc@{}}
\toprule
Model & Text draws & CacheBack draws & Gain (pp) & 95\% CI (pp) \\
\midrule
Qwen 3 8B
& 42 / 40 / 40
& 56 / 52 / 58
& +14.7
& [1.3, 28.0] \\
Ministral 3 14B
& 36 / 36 / 44
& 62 / 56 / 60
& +20.7
& [8.7, 32.7] \\
Gemma 4 12B
& 54 / 50 / 52
& 58 / 56 / 64
& +7.3
& [-3.3, 18.0] \\
Nemotron Nano 2 12B
& 44 / 34 / 38
& 48 / 46 / 56
& +11.3
& [-0.7, 24.0] \\
\midrule
\textbf{Macro-average}
& 44 / 40 / 44
& 56 / 53 / 60
& +13.5
& [6.5, 20.3] \\
\bottomrule
\end{tabular*}
\par
{\raggedright\normalsize
CacheBack uses $4\times$, $32\times$, $4\times$, and $8\times$ for Qwen 3, Ministral 3, Gemma 4, and Nemotron. Family intervals resample questions and receiver draws; the macro-average gives each family equal weight. The macro CI resamples paired questions jointly across families after averaging receiver draws.
\par}
\end{table}

\FloatBarrier

\subsection{LongBench}
\label{app:longbench-draws}

\begin{figure}[htbp]
\centering
\includegraphics[width=0.8\linewidth]{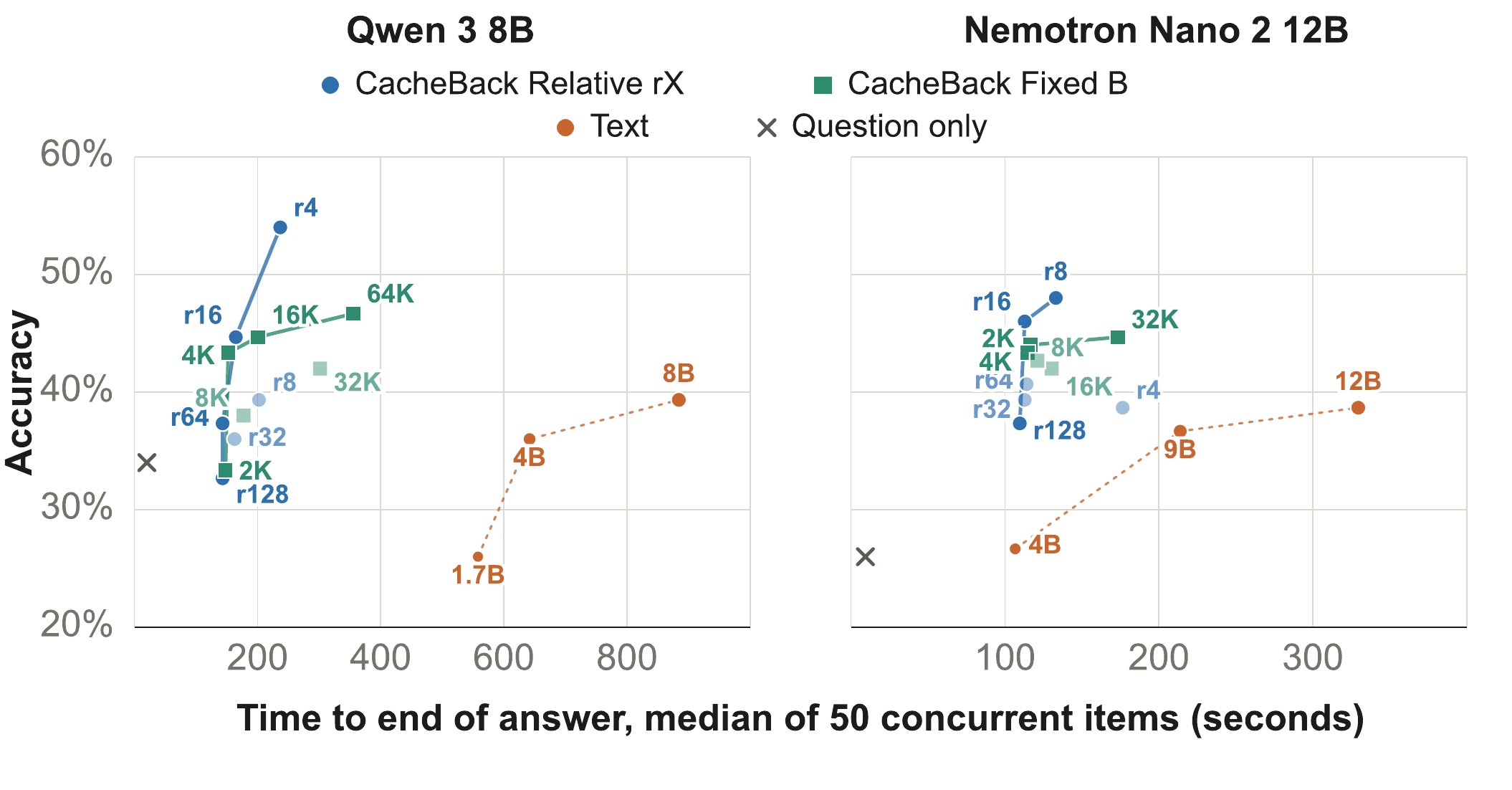}
\caption{LongBench~v2 accuracy and median TTEOA for all evaluated settings. $rX$ denotes relative compression $X$; $x$K denotes a fixed budget of $x$ thousand positions.}
\label{fig:longbench-frontiers}
\end{figure}

Figure~\ref{fig:longbench-frontiers} shows the full accuracy--latency curves for LongBench~v2 Easy. Both relative and fixed CacheBack budgets include settings
with higher accuracy and lower latency than same-size text for Qwen and Nemotron.

\FloatBarrier

\begin{table}[h]
\centering
\caption{LongBench~v2 receiver sampling at selected settings.}
\label{tab:longbench-sampling}
\normalsize
\setlength{\tabcolsep}{4pt}
\renewcommand{\arraystretch}{1}

\begin{tabular*}{\linewidth}{@{\extracolsep{\fill}}llrrr@{}}
\toprule
Model & Setting & Text draws & CacheBack draws & Gain (pp) [95\% CI] \\
\midrule
Qwen 3 8B
& Relative $4\times$
& 36 / 42 / 40
& 56 / 54 / 52
& +14.7 [0.0, 29.3] \\
& Fixed 64K
& 36 / 42 / 40
& 50 / 44 / 46
& +7.3 [-8.0, 22.7] \\
Nemotron Nano 2 12B
& Relative $8\times$
& 40 / 38 / 38
& 44 / 48 / 52
& +9.3 [-5.3, 23.4] \\
& Fixed 32K
& 40 / 38 / 38
& 38 / 46 / 50
& +6.0 [-8.7, 20.7] \\
\bottomrule
\end{tabular*}
\end{table}

Table~\ref{tab:longbench-sampling} shows the three receiver draws at selected settings. We use the same sampling and hierarchical-bootstrap procedure as Appendix~\ref{app:fanoutqa-draws}, holding each message fixed across draws. The selected CacheBack settings improve mean accuracy over same-size text by 6.0--14.7 percentage points.
\FloatBarrier

\subsection{LongBench latency breakdown}
\label{app:longbench-timing}
\begin{figure}[h]
\centering
\includegraphics[width=\linewidth]{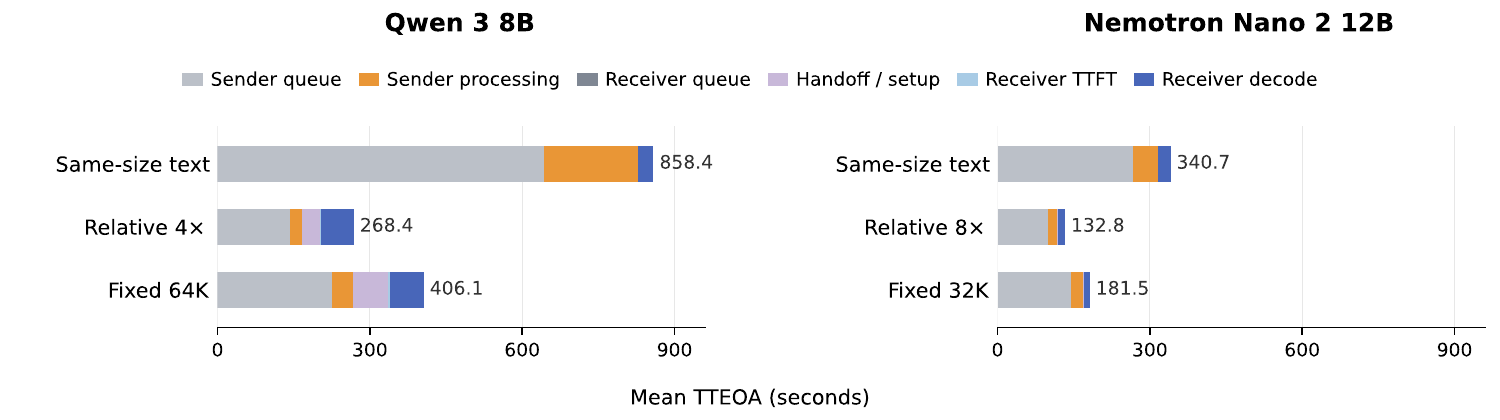}
\caption{LongBench~v2 mean TTEOA by stage at selected settings.}
\label{fig:longbench-latency-breakdown}
\end{figure}

Figure~\ref{fig:longbench-latency-breakdown} decomposes mean TTEOA by stage. CacheBack mainly reduces sender queueing and processing, with the relative budget faster than the fixed budget.

\FloatBarrier
\clearpage

\begin{table}[t]
\centering
\definecolor{longbenchTextBand}{HTML}{F8ECE6}
\definecolor{longbenchLatentBand}{HTML}{E6EDF5}
\caption{LongBench~v2 results on the 50-question Easy panel.}
\label{tab:longbench-full-results}
\normalsize
\setlength{\tabcolsep}{2pt}
\renewcommand{\arraystretch}{1}

\begin{tabular*}{\linewidth}{@{\extracolsep{\fill}}l*{8}{r}@{}}
\toprule
& \multicolumn{4}{c}{\textbf{Qwen 3 8B}}
& \multicolumn{4}{c}{\textbf{Nemotron Nano 2 12B}} \\
\cmidrule(lr){2-5}\cmidrule(lr){6-9}
& \textbf{Accuracy} $\uparrow$
& \multicolumn{3}{c}{\textbf{TTEOA}}
& \textbf{Accuracy} $\uparrow$
& \multicolumn{3}{c}{\textbf{TTEOA}} \\
\textbf{Setting}
& (\%) & p50 (s) & p95 (s) & Speedup
& (\%) & p50 (s) & p95 (s) & Speedup \\
\midrule

Question only
& 34 & 20 & 74 & --
& 26 & 9 & 20 & -- \\

\multicolumn{5}{@{}l}{
\cellcolor{longbenchTextBand}\strut\textbf{Text}}
& \multicolumn{4}{r@{}}{
\cellcolor{longbenchTextBand}\strut} \\

Small
& 26 & 558 & 1014 & 1.6$\times$
& 27 & 107 & 434 & 3.1$\times$ \\

Medium
& 36 & 642 & 1137 & 1.4$\times$
& 37 & 214 & 409 & 1.5$\times$ \\

Same-size
& 39 & 884 & 1627 & 1.0$\times$
& 39 & 329 & 693 & 1.0$\times$ \\

\multicolumn{5}{@{}l}{
\cellcolor{longbenchLatentBand}\strut\textbf{CacheBack, Relative}}
& \multicolumn{4}{r@{}}{
\cellcolor{longbenchLatentBand}\strut} \\

$4\times$
& 54 & 237 & 527 & 3.7$\times$
& 39 & 176 & 288 & 1.9$\times$ \\

$8\times$
& 39 & 202 & 404 & 4.4$\times$
& 48 & 133 & 228 & 2.5$\times$ \\

$16\times$
& 45 & 165 & 276 & 5.4$\times$
& 46 & 113 & 208 & 2.9$\times$ \\

$32\times$
& 36 & 163 & 292 & 5.4$\times$
& 39 & 113 & 199 & 2.9$\times$ \\

$64\times$
& 37 & 143 & 247 & 6.2$\times$
& 41 & 114 & 192 & 2.9$\times$ \\

$128\times$
& 33 & 143 & 245 & 6.2$\times$
& 37 & 110 & 187 & 3.0$\times$ \\

\multicolumn{5}{@{}l}{
\cellcolor{longbenchLatentBand}\strut\textbf{CacheBack, Fixed}}
& \multicolumn{4}{r@{}}{
\cellcolor{longbenchLatentBand}\strut} \\

64K
& 47 & 355 & 801 & 2.5$\times$
& \multicolumn{4}{c}{Exceeds context limit*} \\

32K
& 42 & 302 & 486 & 2.9$\times$
& 45 & 173 & 306 & 1.9$\times$ \\

16K
& 45 & 201 & 414 & 4.4$\times$
& 42 & 130 & 242 & 2.5$\times$ \\

8K
& 38 & 177 & 337 & 5.0$\times$
& 43 & 121 & 218 & 2.7$\times$ \\

4K
& 43 & 152 & 288 & 5.8$\times$
& 43 & 115 & 206 & 2.9$\times$ \\

2K
& 33 & 148 & 274 & 6.0$\times$
& 44 & 117 & 200 & 2.8$\times$ \\

\bottomrule
\end{tabular*}

\par\vspace{5pt}
{\raggedright\normalsize
Text senders are Qwen 3 1.7B/4B/8B and Nemotron 4B/9B/12B; CacheBack senders match the receiver.
Relative uses the listed compression ratio; Fixed caps each message at the listed number of positions.
Each sender reads one document quarter.

TTEOA measures submission to the first receiver answer with all 50 questions submitted together.
The p50 and p95 are across questions.
Speedup is compared to same-size p50. \par
* For Nemotron, the 64K carried budget plus the next document chunk and prompt exceeds the 131K context window on some questions.
\par}
\end{table}

\section{Serving and evaluation settings}
\label{app:serving-settings}

We use the same question panel within each benchmark to compare receiver conditioning across model families and communication topologies. Each run uses one node with eight H100 80GB GPUs and BF16 weights. Models use thinking mode, top-\(p=0.95\), and their recommended sampling settings, with latent vectors rescaled to the mean token-embedding norm. Across latent compression ratios, text sender sizes, two benchmarks, four model families, and three receiver draws, this work used approximately 500 eight-H100 node-hours.

\begin{table}[h]
\centering
\caption{Model and serving settings.}
\label{tab:model-serving-settings}
\normalsize
\renewcommand{\arraystretch}{1}
\setlength{\tabcolsep}{4pt}

\begin{tabularx}{\linewidth}{@{}l*{4}{>{\centering\arraybackslash}X}@{}}
\toprule
& \textbf{Qwen 3} & \textbf{Ministral 3} & \textbf{Gemma 4} & \textbf{Nemotron} \\
& 8B & 14B & 12B & Nano 2 12B \\
\midrule
Global / total layers & 36/36 & 40/40 & 8/48 & 6/62 \\
vLLM & 0.11.1 & 0.26.0 & 0.26.0 & 0.26.0 \\
Receiver window & 131,072 & 180,224 & 262,144 & 131,072 \\
Temperature & 0.6 & 0.7 & 1.0 & 0.6$^{\dagger}$ \\
Top-$k$ & 20 & Off & 64 & Off \\
Alignment map & Off & Off & Off & On \\
Position scaling & YaRN $4\times$ & Native & Native & Native \\
\bottomrule
\end{tabularx}

\par\vspace{3pt}
{\raggedright\normalsize
$^{\dagger}$Nemotron's 4B text sender uses temperature 1.0.
\par}
\end{table}

\begin{table}[h]
\centering
\caption{GPU allocation by communication setting.}
\label{tab:gpu-allocation}
\normalsize
\renewcommand{\arraystretch}{1}
\setlength{\tabcolsep}{4pt}

\begin{tabularx}{\linewidth}{@{}Xrr@{}}
\toprule
\textbf{Setting} & \textbf{Sender GPUs} & \textbf{Receiver GPUs} \\
\midrule
CacheBack, $2\times$ & 4 & 4 \\
CacheBack, $4\times$--$128\times$ & 5 & 3 \\
Text, smaller senders & 6 & 2 \\
Text, same-size senders & \multicolumn{2}{r@{}}{8 fused GPUs} \\
Question only & 0 & 8 \\
Full rows, Gemma and Ministral & 4 & 4 \\
\bottomrule
\end{tabularx}
\end{table}

We choose the sender--receiver GPU split separately for each communication setting because sender and receiver workloads differ. Using the same split would create different queueing bottlenecks across settings. Each receiver admits up to 16 sequences subject to KV capacity. Prefix caching is disabled. Tables~\ref{tab:model-serving-settings} and~\ref{tab:gpu-allocation} give the model settings and GPU allocations. 

\FloatBarrier

\subsection{Timing and output}
\label{app:timing-output}

Time to end of answer (TTEOA) starts when a question is submitted to the node and ends when its first receiver answer finishes. All 50 questions are submitted together, so TTEOA includes queueing for sender and receiver GPUs. We report p50 and p95 over the 50 questions.

We allow receiver generations up to 24{,}000 tokens, including thinking, to avoid truncating long generations. FanOutQA text senders use the same ceiling. On LongBench~v2, text senders use a 16{,}000-token ceiling because some longer generations repeated the same content; Qwen text senders also use a presence penalty of 1.5. Before each run, we verify that the receiver prompt, incoming messages, and 24{,}000-token receiver output ceiling fit within the context window. For Qwen 3 and Nemotron, the Full rows FanOutQA control exceeds that window.

\FloatBarrier

\subsection{FanOutQA evaluation and scoring}
\label{app:answer-scoring}

We evaluate 50 questions from the FanOutQA development split, excluding the questions used for selector development in Section~\ref{sec:selection-behavior}. Each question must provide at least 40,000 tokens of context to every sender under the Qwen 3 tokenizer. We order the remaining questions by the SHA-256 hash of the question ID and take the first 50. We replace one question because source truncation removes a required reference string from the sender context.

We score receiver answers by matching the FanOutQA reference groups. Strict accuracy requires a match for every reference group in the answer after normalization. Loose accuracy is the fraction of matches. We average the three receiver draws for each question, then average across the 50 questions.

Before the final runs, we checked all 498 reference strings against their evidence pages. We corrected nine strings in seven questions: three typos, three values that disagreed with the page, and three answers unsupported by the page. We also match a small set of equivalent forms, including \(72\) and \(72.0\), \texttt{km2} and \(\mathrm{km}^2\), and one shared-surname case that still requires both people. Every model and setting uses the same references and matching rules.

We rescore all 6,900 saved answers from the 46 settings using the original FanOutQA references. CacheBack remains more accurate than same-size text at the reported operating points for all four model families (Table~\ref{tab:fanoutqa-scoring-sensitivity}).

\begin{table}[h]
\centering
\caption{FanOutQA strict accuracy (\%) before and after the reference audit.}
\label{tab:fanoutqa-scoring-sensitivity}

\normalsize
\setlength{\tabcolsep}{4pt}
\renewcommand{\arraystretch}{1}
\begin{tabular*}{\linewidth}{@{\extracolsep{\fill}}lcrrrrrr@{}}
\toprule
& & \multicolumn{3}{c}{Text} & \multicolumn{3}{c}{CacheBack} \\
\cmidrule(lr){3-5}\cmidrule(l){6-8}
Family & Ratio & Before & After & Change
              & Before & After & Change \\
\midrule
Qwen      & $4\times$  & 30.0 & 40.7 & $+10.7$
                        & 41.3 & 55.3 & $+14.0$ \\
Ministral & $32\times$ & 32.0 & 38.7 & $+6.7$
                        & 46.7 & 59.3 & $+12.7$ \\
Gemma     & $4\times$  & 36.7 & 52.0 & $+15.3$
                        & 46.0 & 59.3 & $+13.3$ \\
Nemotron  & $8\times$  & 32.0 & 38.7 & $+6.7$
                        & 38.7 & 50.0 & $+11.3$ \\
\bottomrule
\end{tabular*}
\end{table}

We apply the same matching rules to sender messages. Across the 50 questions, there are 495 audited reference groups. A group counts as present if any of the three sender messages contains an accepted form.

For text, we search only the message sent to the receiver, not the sender's private thinking. For CacheBack, we decode the retained source positions and search the resulting text. A match must come from one sender without crossing discarded positions. We exclude generated latent positions and include prompt framing.

Appendix~\ref{app:evidence-retention} instead measures evidence recall on the selector-development questions.
\newpage

\subsection{LongBench evaluation and scoring}
\label{app:longbench-evaluation}

We evaluate 50 LongBench~v2 questions with documents between 100K and 250K tokens. We use at most one question per document and select the panel by document length and task family using a seeded hash. In an initial
mixed-difficulty pass, the same-size Qwen 3 8B text sender scored 36.2\% on Easy questions and 25.9\% on Hard questions, while 46\% of its reports reached the length ceiling. These reports can end before the sender finishes its notes, making the comparison depend on the report ceiling. For simplicity, we restrict the final panel to Easy questions, add the instruction ``Keep the notes under 8,000 words.'' to the text prompt, and, for Qwen only, use a presence penalty of 1.5. Only 8\% of Qwen 3 8B text reports then reach the ceiling.

Each LongBench~v2 document is split into quarters. The first agent reads the first quarter. Each later agent receives the previous message, reads the next quarter, and sends a new message. The receiver answers the multiple-choice question from the fourth message.

For text, the fourth message is a text message. For CacheBack, it contains the fourth sender's selected positions. The question-only setting answers from the
question and choices alone. Appendix~\ref{app:longbench-prompts} gives the exact prompts.

We score the text after the final \texttt{</think>} tag. We remove asterisks and extract \(A\), \(B\), \(C\), or \(D\) from either \texttt{The correct answer is (X)} or \texttt{The correct answer is X}. The draw scores one if the extracted letter matches the reference answer and zero
otherwise. If neither form appears, it also scores zero. We average the three receiver draws for each question and then average across questions. cross 4,650 receiver generations, 99 required an inserted \texttt{</think>} tag followed by continuation; 76 of these had reached the initial output limit. We allow one continuation of at most 4,096 tokens under the same seed, which lets generation continue into the visible answer.

\definecolor{promptBlue}{HTML}{EFF6FC}
\newsavebox{\appPromptBox}
\newenvironment{appprompt}{%
  \par\medskip\noindent
  \setlength{\fboxsep}{9pt}%
  \begin{lrbox}{\appPromptBox}%
  \begin{minipage}{\dimexpr\linewidth-2\fboxsep\relax}%
  \normalsize\ttfamily\raggedright
}{%
  \end{minipage}\end{lrbox}%
  \colorbox{promptBlue}{\usebox{\appPromptBox}}%
  \par\medskip
}
\clearpage\raggedbottom\section{Prompts and input construction}
\label{app:prompts}
\subsection{FanOutQA prompts}
\label{app:fanoutqa-prompts}

All FanOutQA settings use the same sender evidence and question. Text messages
fill the receiver's report slots; CacheBack supplies continuous rows at the
positions described in Appendix~\ref{app:chat-packing}. The receiver question
and answer instruction are unchanged.

\par\noindent\begin{minipage}{\linewidth}
\noindent{\textbf{Sender}}
\begin{appprompt}
{You are one reasoning worker. Inspect only your private evidence. Work through every relevant name, number, relationship, and uncertainty that could help the coordinator answer the question. Return a complete reasoning handoff, including intermediate reasoning and unresolved alternatives. Do not claim access to any other worker's evidence.}
\par\medskip
{Private evidence:}\par
{\{private evidence\}}
\par\medskip
{Question: \{question\}}
\par\medskip
{Return the complete reasoning handoff now.}
\end{appprompt}
\end{minipage}\par\medskip

\par\noindent\begin{minipage}{\linewidth}
\noindent{\textbf{Receiver}}
\begin{appprompt}
{You are the coordinator. Answer from the supplied handoff and do not invent missing evidence.}
\par\medskip
{Worker reports:}\par
{[worker 0 report]}\par
{\{report 0\}}
\par\medskip
{[worker 1 report]}\par
{\{report 1\}}
\par\medskip
{[worker 2 report]}\par
{\{report 2\}}
\par\medskip
{Question: \{question\}}
\par\medskip
{Give the answer directly and include every requested leaf.}
\end{appprompt}
\end{minipage}\par\medskip

Ministral replaces only the receiver's opening sentence with the following.

\begin{appprompt}
{You are the coordinator. Answer from the supplied handoff; invent no evidence.}
\end{appprompt}

\clearpage\subsection{LongBench prompts}
\label{app:longbench-prompts}

Text and CacheBack use the same question, choices, and final receiver prompt.
Text senders pass messages through the chain, while CacheBack passes selected source and latent positions.

\par\noindent\begin{minipage}{\linewidth}
\noindent{\textbf{Text sender}}
\begin{appprompt}
{You are one reading worker in a chain of 4. You receive the running notes from the earlier parts of a text and one new part. Rewrite the notes into a compact summary of everything read so far that bears on the question below: the names, numbers, dates, events, relationships, and open alternatives it turns on, each stated once in your own words. Do not copy passages, code, tables, or data from the part; state what they say and where they sit. Drop what the question does not need. Keep the notes under 8,000 words. Return only the rewritten notes, not a final answer.}
\par\medskip
{Question and choices:}\par
{\{question\}}\par
{Choices:}\par
{(A) \{choice A\}}\par
{(B) \{choice B\}}\par
{(C) \{choice C\}}\par
{(D) \{choice D\}}
\par\medskip
{Notes from the earlier parts:}\par
{\{running notes\}}
\par\medskip
{Part \{hop\} of 4:}\par
{\{current text chunk\}}
\par\medskip
{Return the rewritten notes now.}
\end{appprompt}
\end{minipage}\par\medskip

\par\noindent\begin{minipage}{\linewidth}
\noindent{\textbf{Final receiver}}
\begin{appprompt}
{The notes a chain of readers kept from the document are below. Answer from them, not from what you already know.}
\par\medskip
{<text>}\par
{\{final notes\}}\par
{</text>}
\par\medskip
{What is the correct answer to this question: \{question\}}\par
{Choices:}\par
{(A) \{choice A\}}\par
{(B) \{choice B\}}\par
{(C) \{choice C\}}\par
{(D) \{choice D\}}
\par\medskip
{Format your response as follows: "The correct answer is (insert answer here)".}
\end{appprompt}
\end{minipage}\par\medskip

\clearpage
\subsection{{Chat-template packing}}
\label{app:chat-packing}

The four model families use different native prompt formats. We preserve each model's format and document where CacheBack rows are inserted. The same packing is used across settings within each benchmark and model family.

For Qwen, CacheBack rows precede the complete rendered prompt. On LongBench, this applies to readers 2--4 and the final receiver; reader 1 has no incoming message.

For Gemma, which we evaluate only on FanOutQA, CacheBack rows precede the complete rendered receiver prompt, including Gemma's native turn framing.

For Nemotron, FanOutQA inserts CacheBack rows into the report slots inside the receiver's user turn. On LongBench, incoming rows precede the complete rendered prompt for readers 2--4. For the final receiver, the rows are inserted inside the user turn after the opening sentence and payload header.

For Ministral, which we also evaluate only on FanOutQA, CacheBack rows precede the complete rendered receiver prompt. Receivers append the following instruction below to the supplied system prompt, while senders use the original system prompt.

\begin{appprompt}
{It is imperative to close the [THINK] tag with a [/THINK] once you are ready to present the answer to the user.}
\end{appprompt}

\end{document}